\documentclass{article}
\usepackage{iclr2027_conference}
\usepackage{times}

\usepackage[english]{babel}
\usepackage{amsmath}
\usepackage{amssymb}
\usepackage{amsthm}
\usepackage{enumitem}

\theoremstyle{definition}
\newtheorem{definition}{Definition}
\newtheorem{fact}{Fact}
\newtheorem{remark}{Remark}
\theoremstyle{plain}
\usepackage{graphicx}
\usepackage[colorlinks=true, allcolors=black]{hyperref}
\usepackage[skins]{tcolorbox}
\usepackage{booktabs}
\usepackage{makecell}
\usepackage{xcolor}
\usepackage[toc,page,header]{appendix}
\usepackage{minitoc}
\definecolor{citeblue}{RGB}{31,60,110}
\hypersetup{citecolor=citeblue}
\definecolor{redmain}{RGB}{163,45,45}
\definecolor{redlight}{RGB}{252,235,235}
\tcbset{
  redbox/.style={enhanced, colback=redlight, colframe=redmain, leftrule=2pt, rightrule=0pt,
    toprule=0pt, bottomrule=0pt, arc=0pt, outer arc=0pt, before upper={\parindent0pt}}}

\title{Divergence controls entropy in distillation}
\author{Nicolas Zucchet and Scott W. Linderman\\Stanford University\\\texttt{\{nzucchet,slinderman\}@stanford.edu}}

\graphicspath{{figures/}}

\iclrpreprint
\begin{document}

\doparttoc
\faketableofcontents

\maketitle

\begin{abstract}
    Distillation has become a core primitive of large language model training, but its properties are not yet well understood.
    We take an entropic perspective, studying how the entropy of the student depends on the data and the divergence that define the distillation objective.
    We prove that forward KL inflates the entropy of the student above that of the teacher.
    Since cross-entropy training is a special case, this yields an identity that we verify quantitatively in pretraining and supervised finetuning.
    Other divergences come with no such guarantee: reverse KL deflates entropy until the gap between student and teacher gets too large, and interpolating between the two changes entropy smoothly early in training but abruptly at convergence.
    The lower entropy of on-policy distillation comes from token-level reverse KL, not from on-policy sampling.
    The divergence therefore acts as an implicit entropy regularizer, whose role is clearest in self-distillation: as conditioning on privileged information deflates entropy, the divergence hyperparameters that work best are those that compensate for it.
\end{abstract}

\vspace{-0.3cm}
\section{Introduction}

Distillation consists of training a student model to mimic the behavior of a teacher model and compress its knowledge \citep{hinton_distilling_2015}.
It is used in increasingly diverse contexts for large language models, from pretraining lightweight, more efficient models \citep{muralidharan2024compact, gemma3team2025}, to solidifying reasoning abilities in posttraining \citep{ma_mopd_2026, deepseek-ai_deepseek-v4_2026}.
In practice, distillation amounts to minimizing a divergence between two categorical probability distributions.
While divergence minimization has been extensively studied elsewhere \citep{minka2005divergence, bishop2006pattern, amari2016information}, its properties are not yet well understood in the context of large language models.

In this work, we study the properties of these algorithms from an entropic perspective, investigating how design choices affect the entropy of the student.
Entropy is fundamentally a measure of uncertainty, whether in thermodynamics \citep{boltzmann1877beziehung} or information theory \citep{shannon1948mathematical}, and it is particularly important for large language models as it relates to the diversity of the output distribution \citep{kirk2024understanding, farquhar2024detecting} and controls exploration in reinforcement learning \citep{cui2025entropy, cheng2026reasoning, wang_beyond_2025}.
It is also easy to compute and log during any training run.
Here, we develop intuition for how distillation shapes entropy.

Our contributions are as follows:
\vspace{-0.2cm}
\begin{itemize}[leftmargin=20pt]
    \item[--] \textbf{Decoupling sampling and divergence choice (Section~\ref{sec:distillation}).} We review distillation, decoupling the choice of the data generation process from that of the divergence, which enables us to study their effects on the student's entropy independently.
    \item[--] \textbf{From divergence to entropy (Section~\ref{sec:entropy}).} In a toy model of language modeling, we show that: forward KL inflates the student's entropy above the teacher's by the residual divergence; reverse KL deflates it until mimicking the teacher gets too difficult and then suddenly inflates it; and interpolating between the two varies entropy smoothly early in training but abruptly at convergence.
    Top-$k$ restrictions also shift entropy, depending on how the tail is handled.
    \item[--] \textbf{At scale, the divergence implicitly controls entropy (Section~\ref{sec:experiments}).} After pretraining and supervised finetuning, the entropy of a language model matches the cross-entropy loss it was trained to minimize; larger models therefore have lower entropy.
    When distilling from a larger teacher, the divergence drives entropy more than on- vs.\ off-policy sampling; in self-distillation, it counters the entropy drop that conditioning on privileged information induces. 
\end{itemize}

\newpage

\section{Decoupling sampling and divergence choice in distillation}
\label{sec:distillation}

\begin{figure}[t]
    \centering
    \includegraphics{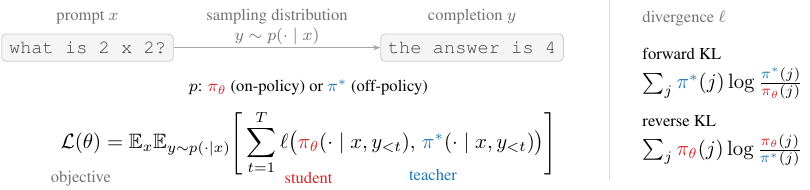}
    \vspace{-0.3cm}
    \caption{\textbf{Decoupling divergence from sampling in distillation.}
    A prompt $x$ is completed by the sampling distribution $p$, and the student $\pi_\theta$ is then compared to the teacher $\pi^*$ on each token of the completion through a divergence $\ell$ between their next-token distributions.
    Writing the objective at the token level keeps the two choices independent, whereas a sequence-level divergence ties them together.
    Details in Section~\ref{sec:distillation}.}
    \label{fig:setup}
\end{figure}

Distillation algorithms, in particular on-policy ones, are often derived by minimizing a divergence between the student's and the teacher's distributions over sequences of tokens.
As we will argue, this couples the choice of token-level objective with the sampling distribution, which makes their respective effects on the student hard to disentangle and can yield suboptimal configurations.
Following \citet{czarnecki2019distilling} and \citet{agarwal2024onpolicy} \textit{inter alia}, we decouple the two, as summarized in Figure~\ref{fig:setup} and detailed in Appendix~\ref{app:derivations}.

\paragraph{Student, teacher and sampling distributions.} 
We consider data consisting of prompts $x$ and completions $y$ generated by a sampling distribution $p$.
On these completions, the student distribution $\pi_\theta(y \mid x)$ parametrized by $\theta$ is compared to that of the teacher $\pi^*(y \mid x)$.
The sampling distribution $p$ can either be the student itself, in which case learning is on-policy, or the teacher, in which case we call it off-policy distillation.
The teacher can either be explicit or implicit, as when training on internet text through next-token cross-entropy minimization.

\paragraph{Distillation objective.}
The objective of distillation is to minimize the expected divergence between the student and the teacher over a dataset of prompts $x$ and completions $y$,
\begin{equation}
    \label{eqn:objective}
    \mathcal{L}(\theta) = \mathbb{E}_x \mathbb{E}_{y \sim p(\cdot \mid x)} \left [ \sum_{t=1}^T \ell \left (\pi_\theta(\cdot \mid x, y_{< t}), \pi^*(\cdot \mid x, y_{< t}) \right ) \right ]\!,
\end{equation}
with $T$ the length of the completion and $\ell$ a divergence between the full next-token distributions of the two models, such as the forward or the reverse KL.
The objective is here defined at the token level, which is what enables us to decouple the sampling distribution $p$ from the divergence choice $\ell$.
Otherwise, the sampling distribution would be implicitly defined by the divergence choice---a sequence-level reverse KL forces on-policy data and a sequence-level forward KL forces off-policy data.
In practice, gradients are typically not backpropagated through the sampling process, even when it is the student itself, with rare exceptions \citep[e.g.,][]{gu2024minillm}.
The same approximation is used when improving the reasoning abilities of language models through reinforcement learning with verifiable rewards \citep{lambert2024tulu3, guo2025deepseekr1}.
It amounts to ignoring the score function term, which biases the gradient but considerably reduces its variance \citep{williams1992simple, roeder2017sticking, naesseth_variational_2018, fu2026revisiting}.
This observation motivates ignoring sequence modeling aspects of distillation in our toy model of Section~\ref{subsec:toy_model}.

\paragraph{Entropy.} Core to our analysis is the average entropy of the model $\pi$, which we define as
\begin{equation}
    \label{eqn:entropy}
    \bar H(\pi) = \mathbb{E}_x\mathbb{E}_{y \sim p( \cdot \mid x)} \left [ \frac{1}{T} \sum_{t=1}^T H \left ( \pi(\cdot \mid x, y_{< t}) \right ) \right ]\!,
\end{equation}
with $H$ the Shannon entropy of a distribution over the vocabulary.
The distribution $\pi$ need not be the one that generates the contexts $(x, y_{< t})$.
When $p = \pi = \pi_\theta$, this is the student's normalized sequence-level entropy.

\section{Properties of divergences and their effect on entropy}
\label{sec:entropy}

\begin{figure}[t]
    \centering
    \includegraphics{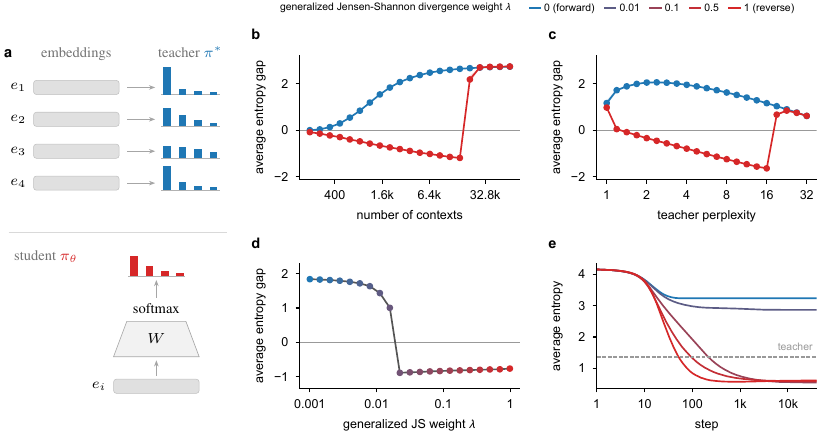}
    \vspace{-0.6cm}
    \caption{\textbf{A toy language model reveals phase transitions in how the divergence affects entropy.}
    \textbf{Toy model (a).} A linear softmax student maps a fixed embedding $e_i$ per context $i$ to next-token probabilities and is trained towards a teacher distribution per context with forward KL, reverse KL, or the generalized Jensen--Shannon (JS) divergence whose weight $\lambda$ interpolates between them (Section~\ref{subsec:toy_model}).
    Its shared weights $W$ prevent it from reproducing all teacher distributions when there are too many contexts.
    \textbf{Simulations (b--e).}
    At convergence, forward KL (blue) always inflates entropy relative to the teacher and reverse KL (red) tends to deflate it, unless the teacher is too hard to model because there are too many contexts (b) or because it is too diffuse (c), in which case reverse KL inflates entropy too.
    Along the generalized JS family (colors in between), entropy smoothly decreases with the weight $\lambda$ early in training (e), whereas at convergence it drops abruptly (d).}
    \label{fig:toy_summary}
\end{figure}

The formalism of the previous section enables us to study the effect of the sampling distribution and the divergence choice independently.
In this section, we restrict ourselves to that of the divergence as it is more tractable to analyze theoretically, and defer the empirical study of the sampling distribution to the next section.
We first introduce a toy model that captures part of language modeling and use it as an illustrative example to highlight the properties of different divergence functions.

\subsection{A toy model of language modeling}
\label{subsec:toy_model}

The canonical example to illustrate the properties of forward and reverse KL divergences is fitting a mixture of Gaussians with a single Gaussian distribution \citep{bishop2006pattern}.
While this has been used to illustrate some of the properties of distillation or reinforcement learning in the context of large language models \citep{agarwal2024onpolicy, gu2024minillm, chen2025retaining, nicolicioiu2026diversity}, its continuous and non-neural-network nature makes it an imperfect proxy for what happens in practice.

Instead, we propose the following toy model, which takes as input a random embedding $e_i$ associated with each context $i$ in the data distribution and outputs a next-token distribution through
\begin{equation}
    \pi_\theta(\cdot \mid i) = \mathrm{softmax}(W e_i).
\end{equation}
Such a model can be understood as the prediction head of a language model whose preceding sequence model provides random embeddings, as an associative memory model \citep{cabannes_scaling_2024,nichani2025factual, zucchet2025facts}, or as the simplest neural network that can produce categorical distributions over possible next tokens.
This model has been shown to be able to fit a number of contexts proportional to the number of parameters of the linear layer \citep{cabannes_scaling_2024, nichani2025factual}.
Varying the number of contexts therefore allows us to control the difficulty of the task, or equivalently the capacity of an implicit teacher.
Another knob is the entropy of the teacher distribution, or equivalently the teacher perplexity $\exp(H(\pi^*))$, that is the effective number of tokens over which the teacher places probability mass; increasing it also makes the task harder \citep{zucchet2026curse}.

Contrary to language models, this toy model has no sequence modeling component, which makes it easier to analyze theoretically.
Little is lost since gradients are not propagated through the sampling process: in both cases, the distillation objective mostly aims at improving next-token prediction and not at shaping the sequence sampling distribution directly.

Throughout this section, we leverage this model when needed to illustrate some of the theoretical properties of the different divergences. 
In our simulations, as a default we set the embedding dimension to $128$, the vocabulary size to $64$, the number of contexts to $2048$, and sample the teacher perplexity log-uniformly between $1$ and $16$.
The model is trained with full-batch Adam \citep{kingma_adam_2015} and a learning rate of $0.1$ for $40$k steps; Appendix~\ref{app:toy-setup} details the setup.

\subsection{Divergence choice only matters when student and teacher differ}
\label{sec:realizable}

Before characterizing what each divergence does to entropy, it is worth delimiting when the choice matters at all.
Indeed, a well-established property of KL divergences is that the geometries of the forward and the reverse KL are equivalent around optimality \citep{amari2016information}.
In our case, this means that if the student is capable of reproducing the teacher and already starts close to it, the learning dynamics of distillation with forward and reverse KL are the same.
Both objectives also share the same global optimum $\pi_\theta = \pi^*$.
More formally, when $\pi_\theta$ is close to $\pi^*$ on all contexts $i$, we have
\begin{equation}
    \nabla_\theta \, \mathbb{E}_i \left [ D_\mathrm{KL}(\pi^*(\cdot \mid i) \parallel \pi_\theta(\cdot \mid i)) \right ] \approx \nabla_\theta \, \mathbb{E}_i \left [ D_\mathrm{KL}(\pi_\theta(\cdot \mid i) \parallel \pi^*(\cdot \mid i)) \right ]
\end{equation}
up to first order.
The student's entropy is therefore also the same along the training trajectory under either objective.
We emphasize that the statement is independent of the sampling distribution $p$, as long as it stays constant during training.
This argument applies more broadly to any $f$-divergence, and in particular to the generalized Jensen--Shannon family that we use later.
A proof of the general statement can be found in Appendix~\ref{app:equivalence}.

A consequence of that result is that any difference in behavior induced by the choice of divergence occurs when the student and the teacher distributions differ significantly, e.g., because the student cannot represent the teacher or because the two start far from each other.
The rest of this section will implicitly assume that we are in such a regime.
Figure~\ref{fig:toy_summary}b illustrates part of this property in our toy model: for a small enough number of contexts, the student and the teacher have almost identical entropy at convergence regardless of the divergence choice, as the student is close to matching the teacher.

\subsection{Forward KL predictably inflates entropy}
\label{sec:forward-kl}

The forward KL divergence is known to be mode-covering \citep{bishop2006pattern, murphy2022probabilistic}: as
\begin{equation}
    D_\mathrm{KL}(\pi^*(\cdot \mid i) \parallel \pi_\theta(\cdot \mid i)) = \sum_j \pi^*(j \mid i) \log \frac{\pi^*(j \mid i)}{\pi_\theta(j \mid i)}
\end{equation}
the objective diverges whenever $\pi^*(j \mid i) > 0$ and $\pi_\theta(j \mid i) \rightarrow 0$, so the student tends to cover all of the teacher's support.
From this intuition, we expect the student to have higher entropy than the teacher; we now show that this does indeed hold at optimality.

Take a neural network with a linear softmax head.
This includes standard language model heads as well as our toy model of Section~\ref{subsec:toy_model}.
At optimality, if we were to multiply the weight matrix by a scalar factor, the loss would not change up to first order.
The consequence of that, which we prove in Appendix~\ref{app:forward-entropy} following classical arguments from information geometry \citep{amari2016information}, is that
\begin{equation}
    \label{eqn:forward-kl-entropy}
    \bar H (\pi_{\theta^*_\mathrm{for}})  = \bar H (\pi^*) + \mathbb{E}_i \left [ D_\mathrm{KL}(\pi^* (\cdot \mid i) \parallel \pi_{\theta^*_\mathrm{for}}(\cdot \mid i)) \right ]
\end{equation}
when $\theta^*_\mathrm{for}$ is a local optimum of the forward KL distillation objective.
In particular, the student necessarily has higher entropy than the teacher on average, which Figure~\ref{fig:toy_summary}b and~\ref{fig:toy_summary}c confirm.
This statement is only relative to the teacher: the forward KL can still deflate entropy over the course of learning if the teacher's entropy is much lower, as Figure~\ref{fig:toy_summary}e illustrates.

Given that the cross-entropy loss minimization objective can be seen as off-policy distillation with the forward KL, in which continuations of the data are samples from an implicit teacher, this result extends to pretraining and supervised finetuning as well.
Expanding the KL, Equation~\ref{eqn:forward-kl-entropy} states that the average entropy of the student equals its average cross-entropy with the teacher, that is its training loss, both measured on the training data.
This does not require the two distributions to be close. 
It only requires a softmax head, as in large language models, and a student at a stationary point of the average forward KL.
A direct consequence is that, among models trained to convergence, a higher loss comes with a higher entropy: a student that cannot match its teacher ends up more uncertain rather than overconfident.
The experiments of Section~\ref{subsec:pretraining} will confirm that these statements hold quantitatively in practice.

\subsection{Reverse KL deflates entropy until the gap gets too large}
\label{sec:reverse-kl}

Whereas the forward KL is mode-covering, the reverse KL is mode-seeking \citep{mackay2003information, bishop2006pattern, margossian2023shrinkage}.
This mode-seeking bias of the reverse KL divergence suggests that a student trained with it would have lower entropy than the teacher.
In practice, this is indeed the case in a wide range of scenarios, but breaks when the task becomes too hard, as illustrated in Figure~\ref{fig:toy_summary}b and~\ref{fig:toy_summary}c when there are too many contexts to model or when the teacher is too diffuse.

Theoretical analysis is not as easy as for the forward KL, so we focus on providing intuition for why this happens.\footnote{The stationarity argument behind Equation~\ref{eqn:forward-kl-entropy} also applies to the reverse KL, but it then constrains the spread of the log probabilities of the student rather than its entropy (Appendix~\ref{app:reverse-flatness}).}
Let us start by rewriting the reverse KL divergence as
\begin{equation}
    D_\mathrm{KL}(\pi_\theta(\cdot \mid i) \parallel \pi^*(\cdot \mid i)) = - \sum_j \pi_\theta(j \mid i) \log \pi^*(j \mid i) - H(\pi_\theta(\cdot \mid i)).
\end{equation}
The first term is the mode-seeking part of the divergence: whenever $\pi^*(j \mid i) \rightarrow 0$, the student must place low probability mass on token $j$ to avoid a large loss.
As the teacher gets more diffuse, the first term flattens out, and the second one, which pushes the student's entropy up, dominates the objective.
As a result, the mode-seeking bias of the reverse KL disappears and the student ends up inflating its entropy.
We can formalize that intuition by analyzing the behavior of the reverse KL divergence in the limit of a flat teacher distribution.
This analysis, provided in Appendix~\ref{app:large-load-uniform}, shows that the reverse KL divergence, as well as any divergence in the generalized Jensen--Shannon family, becomes equivalent to the forward KL divergence in that limit, and therefore inflates entropy.
It does not predict the exact point at which the transition occurs, but it does quantify how the entropy gap between the teacher and the student evolves: the gap widens as the number of contexts grows and narrows as the teacher gets more diffuse, matching Figure~\ref{fig:toy_summary}b and~\ref{fig:toy_summary}c.

One may wonder whether the inflation part of the statement is relevant in practice: in the toy experiments it only occurs when the number of contexts is large or the teacher's output diversity is high, cases in which distillation is unlikely to bring significant performance improvements.
In our large language model experiments of Section~\ref{sec:experiments}, we do not observe any average entropy inflation with the reverse KL.
That said, the analysis here still gives us some insight into the biases of the reverse KL divergence.
Even when inflation does not hold on average, it should still occur on a subset of contexts, and the intuition developed here hints at which ones are more likely to see the entropy of their next-token distribution inflated.

\subsection{In between forward and reverse KL, entropy can sharply change}

Forward and reverse KL are not the only divergences used in practice, and it is common to interpolate between the two through the generalized Jensen--Shannon divergence \citep{vonwerra2020trl, agarwal2024onpolicy, hubotter2026rlsd, zhao_self-distilled_2026, xing2026trust}.
We wonder whether the entropy of the student smoothly and monotonically increases as we interpolate between the two divergences.
The results, illustrated in Figure~\ref{fig:toy_summary}d and~\ref{fig:toy_summary}e, show that it depends on the stage of training: early in training entropy smoothly decreases as the interpolation weight $\lambda$ increases (Figure~\ref{fig:toy_summary}e), whereas at convergence it drops abruptly (Figure~\ref{fig:toy_summary}d).
That said, the phase transition boundary does evolve smoothly as a function of $\lambda$, cf. Figure~\ref{fig:jsd_heatmaps} in the appendix, and divergences with larger $\lambda$ values need harder tasks to inflate entropy.
As a result, one of the consequences of picking a divergence in the generalized Jensen--Shannon family is an implicit regularization on the entropy of the student, which can be either smooth or sharp depending on the training stage.

\subsection{Top-\texorpdfstring{$k$}{k} restrictions shift entropy depending on how the tail is handled}
\label{subsec:topk}

\begin{figure}[]
    \centering
    \includegraphics{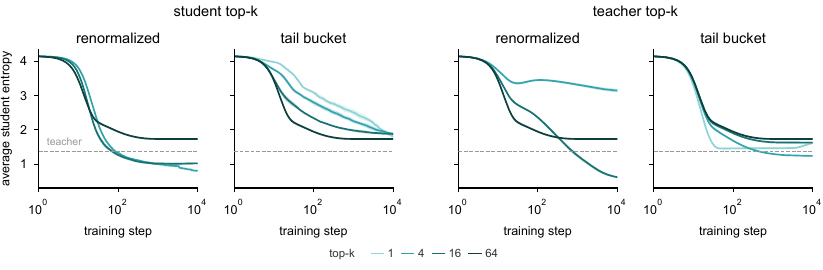}
    \vspace{-0.8cm}
    \caption{\textbf{Restricting the divergence to the top-$k$ tokens shifts entropy depending on how the support and the tail are handled.}
    Same setup as in Figure~\ref{fig:toy_summary} with $512$ contexts, this time varying $k$ out of a vocabulary of $64$, when using the student (left) or the teacher (right) to select the top-$k$ tokens and when using the renormalized or tail bucket cropping strategy.
    Both choices substantially change how the restriction affects entropy.
    Forward KL here; other divergences in Figure~\ref{fig:topk_alpha_seeds}.
    }
    \label{fig:topk_alpha_fkl}
\end{figure}

An important design choice that we have not yet touched upon is how much of the total probability distribution to take into account when computing the divergence.
So far, we have only used full vocabulary divergences, but in practice it is common to restrict the divergence to the top-$k$ tokens of the teacher or student distribution for compute and bandwidth efficiency \citep{fu2026revisiting, hubotter2026rlsd, xing2026trust}.
Two choices arise: which distribution selects the top-$k$ tokens (student or teacher), and how the mass outside the selected set is handled.
The \textit{renormalized} variant renormalizes both distributions over the top-$k$ tokens \citep{fu2026revisiting, xing2026trust}, whereas the \textit{tail bucket} keeps the tail mass as one extra bucket \citep{hubotter2026rlsd, huang2026tail}; see Appendix~\ref{app:topk-definitions}.

We find that top-$k$ restrictions can have a non-negligible effect on the entropy of the student and that the effect depends strongly on the two design choices above.
When the student selects the top-$k$ tokens, renormalization keeps entropy relatively independent of $k$, at least early on, whereas the tail bucket yields increasingly large entropies as $k$ decreases.
When the teacher selects the top-$k$ tokens, the early-training picture reverses.
The results presented here hold for the forward KL divergence, and we additionally sweep over the generalized Jensen--Shannon divergence weight $\lambda$ in Figure~\ref{fig:topk_alpha_seeds} in the appendix, finding that these observations are mostly independent of the choice of divergence, although we observe some interactions.

\section{Entropy in different distillation scenarios}
\label{sec:experiments}

We now turn to empirical validation of the intuition and of the theoretical properties of divergences derived in the previous section, as well as to studying the effect of the sampling distribution.
We consider three settings, which are both practically relevant and will enable us to test different parts of the theory: pretraining, distillation in posttraining for reasoning, and self-distillation.

\subsection{Cross-entropy predicts entropy in pretraining and supervised finetuning}
\label{subsec:pretraining}

\begin{figure}[!t]
    \centering
    \includegraphics{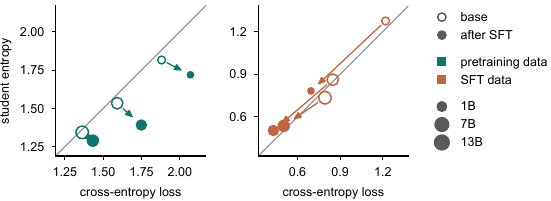}
    \vspace{-0.3cm}
    \caption{\textbf{Cross-entropy tracks entropy in pretraining and supervised finetuning.}
    Average entropy of the student $\bar H(\pi_\theta)$ against its average cross-entropy loss on the data, both computed from the student alone, for OLMo 2 at different sizes (marker size), before (hollow) and after (filled) supervised finetuning (SFT), on the pretraining (left, green) and finetuning (right, orange) mixtures; the grey line is entropy equals cross-entropy.
    The identity holds on the data each model was last trained on, as predicted, and also, without guarantee, on the finetuning data after pretraining, whereas supervised finetuning breaks it on the pretraining data.}
    \label{fig:pretraining_sft}
\end{figure}

Next-token cross-entropy minimization as used during pretraining and supervised finetuning can be understood as sampled off-policy distillation with forward KL, in which the teacher is an implicit distribution over continuations of the data.
In that case the divergence is fixed: we only have access to samples from the teacher distribution, so we cannot use reverse KL or any other divergence that would require direct access to the teacher's probabilities.
As discussed in Section~\ref{sec:forward-kl}, Equation~\ref{eqn:forward-kl-entropy} predicts that the entropy of the student is equal to its cross-entropy loss at convergence, both measured on the training data.
Testing it requires the cross-entropy form of the statement, which only needs samples from the teacher, rather than its form in terms of the entropy of the teacher and the KL divergence to it, which needs the probabilities of the teacher.
Yet, compute-optimal large language models are typically stopped before convergence \citep{kaplan_scaling_2020}, although the trend is to longer pretraining runs \citep{hoffmann_training_2022, sardana_beyond_2024}.
We therefore wonder whether pretraining and supervised finetuning runs are long enough for this relationship to hold; we find it to hold surprisingly precisely in practice, as illustrated in Figure~\ref{fig:pretraining_sft}.
Despite the simplicity and robustness of that result, we are not aware of any prior work that has reported it within the language model literature.

To show this, we run experiments with the OLMo 2 suite of models \citep{olmo_2_2025}, which provides multiple pretraining and supervised finetuned checkpoints with varying model sizes, as well as the pretraining and finetuning datasets that were used to train the models.
For each model size, we evaluate the different checkpoints on both pretraining and finetuning data, and report the entropy as a function of the cross-entropy loss.
Overall, the identity holds, both after pretraining and supervised finetuning.
Cross-entropy loss decreases as we scale models, so larger models have lower entropy.
Figure~\ref{fig:pretraining_pythia_olmo} in the appendix confirms that the same holds for the Pythia suite of models \citep{biderman_pythia_2023}.
Entropy decreases everywhere during finetuning even though we are using a forward KL objective, which inflates entropy.
We reiterate that the inflation is relative to the teacher, and that lower entropy overall reflects the low entropy of the finetuning data, which is for example a well-known property of alignment data \citep{west_base_2025, lanchantin2025divpo, zhang2025verbalized, springer2026annotations}.

Zooming out, this result informs us about how pretrained models fail to capture the data distribution: they tend to leak probability mass outside the support of the data distribution rather than overconfidently predicting high probability tokens.

\subsection{On-policy distillation's lower entropy is due to token-level reverse KL}

\begin{figure}[t]
    \centering
    \includegraphics{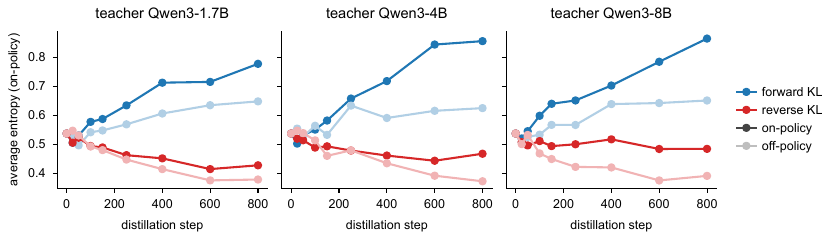}
    \vspace{-0.8cm}
    \caption{\textbf{The lower entropy of on-policy distillation is due to token-level reverse KL, on-policy sampling nudges it up, and teacher size barely matters.}
    Average entropy of the Qwen3-0.6B-Base student on its own MATH500 completions over distillation from Qwen3-Instruct teachers of increasing size in non-thinking mode (left to right).
    Color is the divergence, with the full color for on-policy and a lighter shade for off-policy, and each curve is a single seed.}
    \label{fig:teacher-dist}
\end{figure}

So far, we have focused on the divergence and kept the sampling distribution fixed, which is the setting in which the results of Section~\ref{sec:entropy} hold.
On-policy distillation is increasingly used in the posttraining of large language models \citep{agarwal2024onpolicy, gu2024minillm, yang2025qwen3, thinkingmachines2025onpolicy, xiao2026mimo, yang2026nemotron}, as training on its own completions removes the exposure bias \citep{bengio2015scheduled} that models trained off-policy face.
It is usually paired with reverse KL, as both follow from the same sequence-level derivation \citep{gu2024minillm, thinkingmachines2025onpolicy}, so that a loss of diversity could a priori come from either choice.
In this set of experiments, we ablate the effects of the divergence (forward vs. reverse KL) and the sampling distribution (on-policy vs. off-policy) on entropy.
We distill the Qwen3-Instruct models \citep{yang2025qwen3} in non-thinking mode into a Qwen3-0.6B-Base student on the math problems of DeepScaleR \citep{luo2025deepscaler}.
We measure entropy on MATH500 \citep{lightman2024verify} on student completions in Figure~\ref{fig:teacher-dist}, and on frozen teacher completions in Figure~\ref{fig:teacher-dist-nt} in the appendix.

The divergence has the largest effect on entropy: forward KL raises the entropy of the student and reverse KL lowers it, for every teacher and both sampling distributions, consistently with the ordering of Section~\ref{sec:entropy}.
On-policy distillation consistently yields higher entropy than off-policy distillation, but this effect is two to three times smaller than that of the divergence.
\citet{zhao2026decoupling} similarly report that the prefix source has a much smaller effect on entropy than the direction of the KL, particularly on short completions.
On-policy sampling therefore modulates the effect of the divergence, and the lower entropy of on-policy distillation is mainly due to reverse KL.
Finally, teacher size barely matters: the three teachers have almost the same entropy on their own completions and are all obtained by distillation from the same larger models \citep{yang2025qwen3}, and \citet{li2026rethinking} have shown that teachers trained with the same pipeline induce nearly identical next-token distributions on the contexts the student visits despite their difference in size.

\subsection{Self-distillation deflates entropy, divergence can compensate for it}

In self-distillation, the teacher is the student itself, given additional feedback on how to solve the task.
This sidesteps the need for a more capable model to distill from, and improves performance on a variety of tasks.
\citet{nicolicioiu2026diversity} have shown that conditioning on task feedback decreases the output diversity of the teacher, and thus that of the student.
Here, we focus on how the choice of divergence interacts with this effect.
As the teacher is derived from the student, any entropy the student loses is also lost by the teacher, which can lead to entropy collapse.

We follow the experimental setup of \citet{hubotter2026rlsd} on the chemistry subset of SciKnowEval \citep{feng2024sciknoweval}, which consists of multiple-choice questions testing scientific knowledge and reasoning in chemistry.
We finetune Olmo 3 7B Instruct \citep{olmo2025olmo3}, starting from the default hyperparameters of \citet{hubotter2026rlsd}, and varying three knobs that are important in practice: the divergence within the generalized Jensen--Shannon family, the frequency at which the teacher is updated, and the number of tokens taken into account when computing the divergence.
As in the implementation of \citet{hubotter2026rlsd}, these tokens are the top-$k$ tokens of the student.
The main results are reported in Figure~\ref{fig:self-dist}, and we run the same experiments on GSM8K \citep{cobbe2021gsm8k} in Appendix~\ref{app:self-dist-gsm8k}, with similar conclusions.

Overall, we find that the default hyperparameters are ones that stabilize entropy, and that moving away from them can lead to entropy collapse.
When varying the weight of the generalized Jensen--Shannon divergence (Figure~\ref{fig:self-dist} left), entropy smoothly decreases with $\lambda$, consistently with the early learning dynamics of the toy model (Figure~\ref{fig:toy_summary}e).
In the longer runs of Appendix~\ref{app:self-dist-gsm8k}, entropy sharply increases between $\lambda=0.25$ and $\lambda=0$, although not as abruptly as in the toy model at convergence (Figure~\ref{fig:toy_summary}d).
This could be due to the almost on-policy nature of the data, which makes different runs see different contexts.
With reverse KL, entropy collapses (Figure~\ref{fig:self-dist} left): the entropy deflation of reverse KL compounds with the reduced diversity of the teacher, which in turn lowers the entropy of the next teacher, and so on.
Entropy also collapses when the teacher is the live student (Figure~\ref{fig:self-dist} center), whereas an exponential moving average, even with a decay of $0.95$, behaves like a frozen teacher and prevents the collapse.
Restricting the divergence to fewer tokens (Figure~\ref{fig:self-dist} right) affects entropy as the toy model of Section~\ref{subsec:topk} predicts early in training when the student selects the top-$k$ tokens (Figure~\ref{fig:topk_alpha_fkl} left): with the tail bucket, entropy increases as $k$ decreases, whereas renormalizing leaves it roughly independent of $k$.
Performance-wise, these different hyperparameters account for differences of a few percentage points as long as entropy does not collapse, cf. Figure~\ref{fig:self-dist-accuracy}.

Our results emphasize the importance of separating the choice of divergence from the sampling distribution.
Indeed, deriving self-distillation from a sequence-level reverse KL to justify on-policy sampling forces the choice of reverse KL, which collapses entropy and damages performance.
This is consistent with self-distillation papers \citep[e.g.,][]{shenfeld2026selfdistillation, hubotter2026rlsd, zhao_self-distilled_2026} not finding reverse KL to work best in practice.\footnote{\citet{shenfeld2026selfdistillation} report using reverse KL in the paper, but the code uses forward KL.}

\begin{figure}[t]
    \centering
    \includegraphics{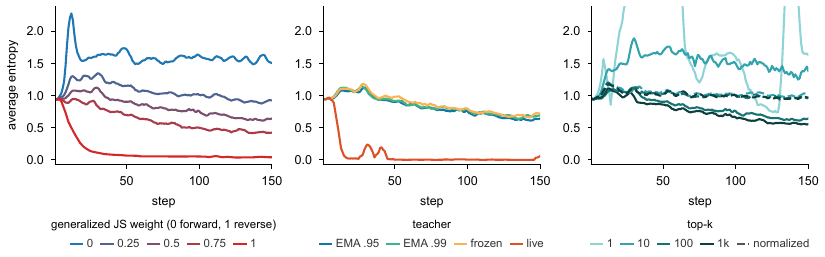}
    \vspace{-0.7cm}
    \caption{
    \textbf{Self-distillation tends to deflate entropy, the choice of divergence must compensate for it to avoid entropy collapse.}
    Self-distillation experiments on the Chemistry dataset following \citet{hubotter2026rlsd}, varying the weight of the generalized Jensen--Shannon (JS) divergence (left), the frequency at which the teacher is updated (center), and how many tokens are taken into account (right); defaults are $\lambda=0.5$, a teacher that is an exponential moving average (EMA) of the student with decay $0.95$, and the top-$100$ tokens of the student with tail bucket.
    Each curve is a single seed.
    Conditioning on privileged information lowers the entropy of the teacher \citep{nicolicioiu2026diversity}, and as the teacher derives from the student, this can compound into entropy collapse, e.g., with reverse KL (left) or a live teacher (center).
    Divergences that inflate entropy, a slowly moving teacher, or a tail bucket (right) counterbalance it.
    }
    \label{fig:self-dist}
\end{figure}

\section{Conclusion}

The divergence used in distillation is often inherited rather than chosen: a sequence-level derivation ties reverse KL to on-policy data and forward KL to off-policy data. 
Once we disentangled these choices, we found that the divergence, more than the sampling distribution, acts as an implicit regularizer on the student.
Forward KL inflates the entropy of the student above that of the teacher by exactly the residual divergence on average, so the entropy of a converged pretraining run can be read from its loss.
Reverse KL deflates entropy, but only as long as the student can mimic the teacher; whether it does is a property of the divergence and the task together, not of the divergence alone.
In between, and depending on how the tail of the distribution is handled, entropy can be tuned, smoothly early in training and abruptly at convergence.
These effects compound in self-distillation, where the teacher is derived from the student: any entropy drop feeds back into the next teacher, and the hyperparameters that work best are those whose inflation counteracts the drop that conditioning on privileged information induces.

Practically, our results suggest choosing the sampling distribution and the divergence separately, and treating the latter as an entropy knob rather than as a consequence of the former.

We obtained these results with a toy model of language modeling, simple enough to admit precise mathematical statements yet rich enough to reproduce key phenomena at scale.
We hope that it will replace the Gaussian mixtures inherited from variational inference as the go-to example to reason about distillation in large language models, and that it will prove useful beyond entropy.

\ificlrfinal
\subsubsection*{Acknowledgments}
N.Z. is supported by Postdoc.Mobility grant P500-2\_235376 from the Swiss National Science Foundation.
S.W.L. is supported by grants from the NIH (U01NS136507, R01NS131987, RF1MH133778, R01AG097491, \& R01NS130789), the NSF (2440859), and the Simons and McKnight Foundations.

The experiments were performed on the Stanford Marlowe computing cluster \citep{kapfer_marlowe_2025}.
We thank Stanford University and the Stanford Research Computing Center for providing computational resources and support that contributed to these research results.
\fi

\subsection*{AI use statement}

In this work, we used generative AI tools for
\begin{itemize}
    \item[--] helping with the literature review.
    \item[--] assisting in the development of the theory by formulating mathematical claims and providing insights into how to prove them.
    We used these tools to draft proofs and carefully rewrote them ourselves to ensure correctness and improve clarity.
    \item[--] providing feedback on experiments, writing code, running experiments, analyzing the results and producing the figures.
    \item[--] polishing the writing of the paper and drafting some sections of the appendix.
    The first draft of the main text was written exclusively by the authors.
\end{itemize}
We take responsibility for the final content of this work, including text, claims or artifacts produced with the aid of generative AI.

\subsection*{Reproducibility statement}

The code to reproduce all experiments and figures, together with the data behind every figure, is available at \url{https://github.com/NicolasZucchet/Entropy-in-distillation}.

{
\small
\bibliographystyle{iclr2027_conference}\bibliography{main}

\begin{thebibliography}{70}
\providecommand{\natexlab}[1]{#1}
\providecommand{\url}[1]{\texttt{#1}}
\expandafter\ifx\csname urlstyle\endcsname\relax
  \providecommand{\doi}[1]{doi: #1}\else
  \providecommand{\doi}{doi: \begingroup \urlstyle{rm}\Url}\fi

\bibitem[Agarwal et~al.(2024)Agarwal, Vieillard, Zhou, Stanczyk, Ramos, Geist,
  and Bachem]{agarwal2024onpolicy}
Rishabh Agarwal, Nino Vieillard, Yongchao Zhou, Piotr Stanczyk, Sabela Ramos,
  Matthieu Geist, and Olivier Bachem.
\newblock On-policy distillation of language models: {L}earning from
  self-generated mistakes.
\newblock In \emph{International conference on learning representations}, 2024.

\bibitem[Amari(2016)]{amari2016information}
Shun-ichi Amari.
\newblock \emph{Information geometry and its applications}.
\newblock Springer, 2016.

\bibitem[Balles \& Hennig(2018)Balles and Hennig]{balles_dissecting_2018}
Lukas Balles and Philipp Hennig.
\newblock Dissecting {Adam}: the sign, magnitude and variance of stochastic
  gradients.
\newblock In \emph{International conference on machine learning}, 2018.

\bibitem[Bengio et~al.(2015)Bengio, Vinyals, Jaitly, and
  Shazeer]{bengio2015scheduled}
Samy Bengio, Oriol Vinyals, Navdeep Jaitly, and Noam Shazeer.
\newblock Scheduled sampling for sequence prediction with recurrent neural
  networks.
\newblock In \emph{Advances in neural information processing systems}, 2015.

\bibitem[Biderman et~al.(2023)Biderman, Schoelkopf, Anthony, Bradley,
  O’Brien, Hallahan, Khan, Purohit, Prashanth, Raff, and
  {others}]{biderman_pythia_2023}
Stella Biderman, Hailey Schoelkopf, Quentin~Gregory Anthony, Herbie Bradley,
  Kyle O’Brien, Eric Hallahan, Mohammad~Aflah Khan, Shivanshu Purohit,
  USVSN~Sai Prashanth, Edward Raff, and {others}.
\newblock Pythia: {A} suite for analyzing large language models across training
  and scaling.
\newblock In \emph{International conference on machine learning}, 2023.

\bibitem[Bishop(2006)]{bishop2006pattern}
Christopher~M. Bishop.
\newblock \emph{Pattern recognition and machine learning}.
\newblock Springer, 2006.

\bibitem[B{\l}asiok et~al.(2023)B{\l}asiok, Gopalan, Hu, and
  Nakkiran]{blasiok2023when}
Jaros{\l}aw B{\l}asiok, Parikshit Gopalan, Lunjia Hu, and Preetum Nakkiran.
\newblock When does optimizing a proper loss yield calibration?
\newblock In \emph{Advances in neural information processing systems}, 2023.

\bibitem[Boltzmann(1877)]{boltzmann1877beziehung}
Ludwig Boltzmann.
\newblock {\"{U}ber die Beziehung zwischen dem zweiten Hauptsatze der
  mechanischen W\"{a}rmetheorie und der Wahrscheinlichkeitsrechnung respektive
  den S\"{a}tzen \"{u}ber das W\"{a}rmegleichgewicht}.
\newblock \emph{Sitzungsberichte der Kaiserlichen Akademie der Wissenschaften.
  Mathematisch-Naturwissenschaftliche Classe}, 76, 1877.

\bibitem[Cabannes et~al.(2024)Cabannes, Dohmatob, and
  Bietti]{cabannes_scaling_2024}
Vivien Cabannes, Elvis Dohmatob, and Alberto Bietti.
\newblock Scaling laws for associative memories.
\newblock In \emph{International conference on learning representations}, 2024.

\bibitem[Chen et~al.(2026)Chen, Razin, Narasimhan, and Chen]{chen2025retaining}
Howard Chen, Noam Razin, Karthik Narasimhan, and Danqi Chen.
\newblock Retaining by doing: {T}he role of on-policy data in mitigating
  forgetting.
\newblock In \emph{International conference on machine learning}, 2026.

\bibitem[Cheng et~al.(2026)Cheng, Huang, Zhu, Dai, Zhao, Zhang, and
  Wei]{cheng2026reasoning}
Daixuan Cheng, Shaohan Huang, Xuekai Zhu, Bo~Dai, Xin Zhao, Zhenliang Zhang,
  and Furu Wei.
\newblock Reasoning with exploration: An entropy perspective.
\newblock In \emph{AAAI conference on artificial intelligence}, 2026.

\bibitem[Clart{\'e} et~al.(2023)Clart{\'e}, Loureiro, Krzakala, and
  Zdeborov{\'a}]{clarte2023expectation}
Lucas Clart{\'e}, Bruno Loureiro, Florent Krzakala, and Lenka Zdeborov{\'a}.
\newblock Expectation consistency for calibration of neural networks.
\newblock In \emph{Uncertainty in artificial intelligence}, 2023.

\bibitem[Cobbe et~al.(2021)Cobbe, Kosaraju, Bavarian, Chen, Jun, Kaiser,
  Plappert, Tworek, Hilton, Nakano, Hesse, and Schulman]{cobbe2021gsm8k}
Karl Cobbe, Vineet Kosaraju, Mohammad Bavarian, Mark Chen, Heewoo Jun, Lukasz
  Kaiser, Matthias Plappert, Jerry Tworek, Jacob Hilton, Reiichiro Nakano,
  Christopher Hesse, and John Schulman.
\newblock Training verifiers to solve math word problems.
\newblock \emph{arXiv preprint arXiv:2110.14168}, 2021.

\bibitem[Cui et~al.(2025)Cui, Zhang, Chen, Yuan, Wang, Zuo, Li, Fan, Chen,
  Chen, Liu, Peng, Bai, Ouyang, Cheng, Zhou, and Ding]{cui2025entropy}
Ganqu Cui, Yuchen Zhang, Jiacheng Chen, Lifan Yuan, Zhi Wang, Yuxin Zuo,
  Haozhan Li, Yuchen Fan, Huayu Chen, Weize Chen, Zhiyuan Liu, Hao Peng, Lei
  Bai, Wanli Ouyang, Yu~Cheng, Bowen Zhou, and Ning Ding.
\newblock The entropy mechanism of reinforcement learning for reasoning
  language models.
\newblock \emph{arXiv preprint arXiv:2505.22617}, 2025.

\bibitem[Czarnecki et~al.(2019)Czarnecki, Pascanu, Osindero, Jayakumar,
  {\'S}wirszcz, and Jaderberg]{czarnecki2019distilling}
Wojciech~M. Czarnecki, Razvan Pascanu, Simon Osindero, Siddhant~M. Jayakumar,
  Grzegorz {\'S}wirszcz, and Max Jaderberg.
\newblock Distilling policy distillation.
\newblock In \emph{International conference on artificial intelligence and
  statistics}, 2019.

\bibitem[DeepSeek-AI(2026)]{deepseek-ai_deepseek-v4_2026}
DeepSeek-AI.
\newblock {DeepSeek}-{V4}: towards highly efficient million-token context
  intelligence.
\newblock \emph{arXiv preprint arXiv:2606.19348}, 2026.

\bibitem[Farquhar et~al.(2024)Farquhar, Kossen, Kuhn, and
  Gal]{farquhar2024detecting}
Sebastian Farquhar, Jannik Kossen, Lorenz Kuhn, and Yarin Gal.
\newblock Detecting hallucinations in large language models using semantic
  entropy.
\newblock \emph{Nature}, 630\penalty0 (8017), 2024.

\bibitem[Feng et~al.(2024)Feng, Shen, Wang, Zhuang, Tang, Zhang, and
  Ding]{feng2024sciknoweval}
Kehua Feng, Xinyi Shen, Weijie Wang, Xiang Zhuang, Yuqi Tang, Qiang Zhang, and
  Keyan Ding.
\newblock {SciKnowEval}: evaluating multi-level scientific knowledge of large
  language models.
\newblock \emph{arXiv preprint arXiv:2406.09098}, 2024.

\bibitem[Fu et~al.(2026)Fu, Huang, Jiang, Liu, Jiang, Zhu, and
  Zhao]{fu2026revisiting}
Yuqian Fu, Haohuan Huang, Kaiwen Jiang, Jiacai Liu, Zhuo Jiang, Yuanheng Zhu,
  and Dongbin Zhao.
\newblock Revisiting on-policy distillation: empirical failure modes and simple
  fixes.
\newblock In \emph{Conference on language modeling}, 2026.

\bibitem[Gao et~al.(2020)Gao, Biderman, Black, Golding, Hoppe, Foster, Phang,
  He, Thite, Nabeshima, Presser, and Leahy]{gao2020pile}
Leo Gao, Stella Biderman, Sid Black, Laurence Golding, Travis Hoppe, Charles
  Foster, Jason Phang, Horace He, Anish Thite, Noa Nabeshima, Shawn Presser,
  and Connor Leahy.
\newblock The {Pile}: an {800GB} dataset of diverse text for language modeling.
\newblock \emph{arXiv preprint arXiv:2101.00027}, 2020.

\bibitem[{Gemma Team}(2025)]{gemma3team2025}
{Gemma Team}.
\newblock {Gemma} 3 technical report.
\newblock \emph{arXiv preprint arXiv:2503.19786}, 2025.

\bibitem[Gu et~al.(2024)Gu, Dong, Wei, and Huang]{gu2024minillm}
Yuxian Gu, Li~Dong, Furu Wei, and Minlie Huang.
\newblock {MiniLLM}: knowledge distillation of large language models.
\newblock In \emph{International conference on learning representations}, 2024.

\bibitem[Guo et~al.(2025)Guo, Yang, Zhang, Song, Wang, Zhu, Xu, Zhang, Ma, Bi,
  et~al.]{guo2025deepseekr1}
Daya Guo, Dejian Yang, Haowei Zhang, Junxiao Song, Peiyi Wang, Qihao Zhu,
  Runxin Xu, Ruoyu Zhang, Shirong Ma, Xiao Bi, et~al.
\newblock {DeepSeek-R1} incentivizes reasoning in {LLM}s through reinforcement
  learning.
\newblock \emph{Nature}, 645\penalty0 (8081), 2025.

\bibitem[Hinton et~al.(2015)Hinton, Vinyals, and Dean]{hinton_distilling_2015}
Geoffrey Hinton, Oriol Vinyals, and Jeff Dean.
\newblock Distilling the knowledge in a neural network.
\newblock \emph{arXiv preprint arXiv:1503.02531}, 2015.

\bibitem[Hoffmann et~al.(2022)Hoffmann, Borgeaud, Mensch, Buchatskaya, Cai,
  Rutherford, Casas, Hendricks, Welbl, Clark, and
  {others}]{hoffmann_training_2022}
Jordan Hoffmann, Sebastian Borgeaud, Arthur Mensch, Elena Buchatskaya, Trevor
  Cai, Eliza Rutherford, Diego de~Las Casas, Lisa~Anne Hendricks, Johannes
  Welbl, Aidan Clark, and {others}.
\newblock An empirical analysis of compute-optimal large language model
  training.
\newblock In \emph{Advances in neural information processing systems}, 2022.

\bibitem[Huang \& Wei(2026)Huang and Wei]{huang2026tail}
Huipeng Huang and Hongxin Wei.
\newblock Tail-aware top-$k$ on-policy distillation.
\newblock \emph{arXiv preprint arXiv:2608.14728}, 2026.

\bibitem[H{\"u}botter et~al.(2026)H{\"u}botter, L{\"u}beck, Behric, Baumann,
  Bagatella, Marta, Hakimi, Shenfeld, Kleine~Buening, Guestrin, and
  Krause]{hubotter2026rlsd}
Jonas H{\"u}botter, Frederike L{\"u}beck, Lejs Behric, Anton Baumann, Marco
  Bagatella, Daniel Marta, Ido Hakimi, Idan Shenfeld, Thomas Kleine~Buening,
  Carlos Guestrin, and Andreas Krause.
\newblock Reinforcement learning via self-distillation.
\newblock In \emph{International conference on machine learning}, 2026.

\bibitem[Kapfer et~al.(2025)Kapfer, Stine, Narasimhan, Mentzel, and
  Cand\`es]{kapfer_marlowe_2025}
Craig Kapfer, Kurt Stine, Balasubramanian Narasimhan, Christopher Mentzel, and
  Emmanuel Cand\`es.
\newblock Marlowe: Stanford's {GPU}-based computational instrument.
\newblock Zenodo, 2025.

\bibitem[Kaplan et~al.(2020)Kaplan, McCandlish, Henighan, Brown, Chess, Child,
  Gray, Radford, Wu, and Amodei]{kaplan_scaling_2020}
Jared Kaplan, Sam McCandlish, Tom Henighan, Tom~B Brown, Benjamin Chess, Rewon
  Child, Scott Gray, Alec Radford, Jeffrey Wu, and Dario Amodei.
\newblock Scaling laws for neural language models.
\newblock \emph{arXiv preprint arXiv:2001.08361}, 2020.

\bibitem[Kingma \& Ba(2015)Kingma and Ba]{kingma_adam_2015}
Diederik~P. Kingma and Jimmy Ba.
\newblock Adam: a method for stochastic optimization.
\newblock In \emph{International conference on learning representations}, 2015.

\bibitem[Kirk et~al.(2024)Kirk, Mediratta, Nalmpantis, Luketina, Hambro,
  Grefenstette, and Raileanu]{kirk2024understanding}
Robert Kirk, Ishita Mediratta, Christoforos Nalmpantis, Jelena Luketina, Eric
  Hambro, Edward Grefenstette, and Roberta Raileanu.
\newblock Understanding the effects of {RLHF} on {LLM} generalisation and
  diversity.
\newblock In \emph{International conference on learning representations}, 2024.

\bibitem[Kwon et~al.(2023)Kwon, Li, Zhuang, Sheng, Zheng, Yu, Gonzalez, Zhang,
  and Stoica]{kwon2023vllm}
Woosuk Kwon, Zhuohan Li, Siyuan Zhuang, Ying Sheng, Lianmin Zheng, Cody~Hao Yu,
  Joseph~E. Gonzalez, Hao Zhang, and Ion Stoica.
\newblock Efficient memory management for large language model serving with
  {P}aged{A}ttention.
\newblock In \emph{Symposium on operating systems principles}, 2023.

\bibitem[Lambert et~al.(2025)Lambert, Morrison, Pyatkin, Huang, Ivison,
  Brahman, Miranda, Liu, Dziri, Lyu, Gu, Malik, Graf, Hwang, Yang, Bras,
  Tafjord, Wilhelm, Soldaini, Smith, Wang, Dasigi, and
  Hajishirzi]{lambert2024tulu3}
Nathan Lambert, Jacob Morrison, Valentina Pyatkin, Shengyi Huang, Hamish
  Ivison, Faeze Brahman, Lester James~V. Miranda, Alisa Liu, Nouha Dziri, Shane
  Lyu, Yuling Gu, Saumya Malik, Victoria Graf, Jena~D. Hwang, Jiangjiang Yang,
  Ronan~Le Bras, Oyvind Tafjord, Chris Wilhelm, Luca Soldaini, Noah~A. Smith,
  Yizhong Wang, Pradeep Dasigi, and Hannaneh Hajishirzi.
\newblock {T\"ulu} 3: Pushing frontiers in open language model post-training.
\newblock In \emph{Conference on language modeling}, 2025.

\bibitem[Lanchantin et~al.(2025)Lanchantin, Chen, Dhuliawala, Yu, Weston,
  Sukhbaatar, and Kulikov]{lanchantin2025divpo}
Jack Lanchantin, Angelica Chen, Shehzaad Dhuliawala, Ping Yu, Jason Weston,
  Sainbayar Sukhbaatar, and Ilia Kulikov.
\newblock Diverse preference optimization.
\newblock \emph{arXiv preprint arXiv:2501.18101}, 2025.

\bibitem[Li et~al.(2026)Li, Zuo, He, Zhang, Xiao, Qian, Yu, Gao, Yang, Liu, and
  Ding]{li2026rethinking}
Yaxuan Li, Yuxin Zuo, Bingxiang He, Jinqian Zhang, Chaojun Xiao, Cheng Qian,
  Tianyu Yu, Huan-ang Gao, Wenkai Yang, Zhiyuan Liu, and Ning Ding.
\newblock Rethinking on-policy distillation of large language models:
  {P}henomenology, mechanism, and recipe.
\newblock \emph{arXiv preprint arXiv:2604.13016}, 2026.

\bibitem[Lightman et~al.(2024)Lightman, Kosaraju, Burda, Edwards, Baker, Lee,
  Leike, Schulman, Sutskever, and Cobbe]{lightman2024verify}
Hunter Lightman, Vineet Kosaraju, Yuri Burda, Harrison Edwards, Bowen Baker,
  Teddy Lee, Jan Leike, John Schulman, Ilya Sutskever, and Karl Cobbe.
\newblock Let's verify step by step.
\newblock In \emph{International conference on learning representations}, 2024.

\bibitem[Lu \& {Thinking Machines Lab}(2025)Lu and {Thinking Machines
  Lab}]{thinkingmachines2025onpolicy}
Kevin Lu and {Thinking Machines Lab}.
\newblock On-policy distillation.
\newblock Thinking Machines Lab: Connectionism,
  \url{https://thinkingmachines.ai/blog/on-policy-distillation/}, 2025.

\bibitem[Luo et~al.(2025)Luo, Tan, Wong, Shi, Tang, Roongta, Cai, Luo, Li,
  Popa, and Stoica]{luo2025deepscaler}
Michael Luo, Sijun Tan, Justin Wong, Xiaoxiang Shi, William~Y. Tang, Manan
  Roongta, Colin Cai, Jeffrey Luo, Li~Erran Li, Raluca~Ada Popa, and Ion
  Stoica.
\newblock {DeepScaleR}: {S}urpassing {O1}-preview with a 1.5{B} model by
  scaling {RL}.
\newblock Blog post, 2025.

\bibitem[Ma et~al.(2026)Ma, Wei, Zhao, Zhang, Xiao, Li, Yang, Gao, Wang, Li,
  Dong, Sui, and Luo]{ma_mopd_2026}
Wenhan Ma, Jianyu Wei, Liang Zhao, Hailin Zhang, Bangjun Xiao, Lei Li, Qibin
  Yang, Bofei Gao, Yudong Wang, Rang Li, Jinhao Dong, Zhifang Sui, and Fuli
  Luo.
\newblock {MOPD}: multi-teacher on-policy distillation for capability
  integration in {LLM} post-training.
\newblock \emph{arXiv preprint arXiv:2606.30406}, 2026.

\bibitem[MacKay(2003)]{mackay2003information}
David J.~C. MacKay.
\newblock \emph{Information theory, inference, and learning algorithms}.
\newblock Cambridge University Press, 2003.

\bibitem[Margossian \& Saul(2023)Margossian and Saul]{margossian2023shrinkage}
Charles~C. Margossian and Lawrence~K. Saul.
\newblock The shrinkage-delinkage trade-off: an analysis of factorized
  {G}aussian approximations for variational inference.
\newblock In \emph{Uncertainty in artificial intelligence}, 2023.

\bibitem[Minka(2005)]{minka2005divergence}
Tom Minka.
\newblock Divergence measures and message passing.
\newblock Technical Report MSR-TR-2005-173, Microsoft Research, 2005.

\bibitem[Muralidharan et~al.(2024)Muralidharan, Sreenivas, Joshi, Chochowski,
  Patwary, Shoeybi, Catanzaro, Kautz, and Molchanov]{muralidharan2024compact}
Saurav Muralidharan, Sharath~Turuvekere Sreenivas, Raviraj Joshi, Marcin
  Chochowski, Mostofa Patwary, Mohammad Shoeybi, Bryan Catanzaro, Jan Kautz,
  and Pavlo Molchanov.
\newblock Compact language models via pruning and knowledge distillation.
\newblock In \emph{Advances in neural information processing systems}, 2024.

\bibitem[Murphy(2022)]{murphy2022probabilistic}
Kevin~P. Murphy.
\newblock \emph{Probabilistic machine learning: an introduction}.
\newblock MIT Press, 2022.

\bibitem[Naesseth et~al.(2018)Naesseth, Linderman, Ranganath, and
  Blei]{naesseth_variational_2018}
Christian Naesseth, Scott Linderman, Rajesh Ranganath, and David Blei.
\newblock Variational sequential {Monte} {Carlo}.
\newblock In \emph{International conference on artificial intelligence and
  statistics}, 2018.

\bibitem[Nichani et~al.(2025)Nichani, Lee, and Bietti]{nichani2025factual}
Eshaan Nichani, Jason~D. Lee, and Alberto Bietti.
\newblock Understanding factual recall in transformers via associative
  memories.
\newblock In \emph{International conference on learning representations}, 2025.

\bibitem[Nicolicioiu et~al.(2026)Nicolicioiu, Pezeshki, and
  Courville]{nicolicioiu2026diversity}
Andrei~Liviu Nicolicioiu, Mohammad Pezeshki, and Aaron Courville.
\newblock On-policy self-distillation with sampled demonstrations reduces
  output diversity.
\newblock \emph{arXiv preprint arXiv:2606.26091}, 2026.

\bibitem[Polyanskiy \& Wu(2025)Polyanskiy and Wu]{polyanskiy2025information}
Yury Polyanskiy and Yihong Wu.
\newblock \emph{Information theory: {F}rom coding to learning}.
\newblock Cambridge University Press, 2025.

\bibitem[Roeder et~al.(2017)Roeder, Wu, and Duvenaud]{roeder2017sticking}
Geoffrey Roeder, Yuhuai Wu, and David Duvenaud.
\newblock Sticking the landing: {S}imple, lower-variance gradient estimators
  for variational inference.
\newblock In \emph{Advances in neural information processing systems}, 2017.

\bibitem[Sardana et~al.(2024)Sardana, Portes, Doubov, and
  Frankle]{sardana_beyond_2024}
Nikhil Sardana, Jacob Portes, Sasha Doubov, and Jonathan Frankle.
\newblock Beyond {Chinchilla}-optimal: {Accounting} for inference in language
  model scaling laws.
\newblock In \emph{International conference on machine learning}, 2024.

\bibitem[Schulman et~al.(2016)Schulman, Moritz, Levine, Jordan, and
  Abbeel]{schulman2016gae}
John Schulman, Philipp Moritz, Sergey Levine, Michael~I. Jordan, and Pieter
  Abbeel.
\newblock High-dimensional continuous control using generalized advantage
  estimation.
\newblock In \emph{International conference on learning representations}, 2016.

\bibitem[Shannon(1948)]{shannon1948mathematical}
Claude~Elwood Shannon.
\newblock A mathematical theory of communication.
\newblock \emph{The Bell system technical journal}, 27\penalty0 (3), 1948.

\bibitem[Shenfeld et~al.(2026)Shenfeld, Damani, H{\"u}botter, and
  Agrawal]{shenfeld2026selfdistillation}
Idan Shenfeld, Mehul Damani, Jonas H{\"u}botter, and Pulkit Agrawal.
\newblock Self-distillation enables continual learning.
\newblock In \emph{International conference on machine learning}, 2026.

\bibitem[Springer et~al.(2026)Springer, Advani, Aichberger, Bradley, Malach,
  Saremi, Williamson, Nakkiran, Littwin, and
  Raghunathan]{springer2026annotations}
Jacob~Mitchell Springer, Madhu Advani, Lukas Aichberger, Arwen Bradley, Eran
  Malach, Omid Saremi, Sinead Williamson, Preetum Nakkiran, Etai Littwin, and
  Aditi Raghunathan.
\newblock Annotations mitigate post-training mode collapse.
\newblock In \emph{International conference on machine learning}, 2026.

\bibitem[{Team Olmo}(2025)]{olmo2025olmo3}
{Team Olmo}.
\newblock {Olmo} 3.
\newblock \emph{arXiv preprint arXiv:2512.13961}, 2025.

\bibitem[{Team OLMo}(2025)]{olmo_2_2025}
{Team OLMo}.
\newblock 2 {OLMo} 2 {Furious}.
\newblock In \emph{Conference on language modeling}, 2025.

\bibitem[von Werra et~al.(2020)von Werra, Belkada, Tunstall, Beeching, Thrush,
  Lambert, Huang, Rasul, and Gallou{\'e}dec]{vonwerra2020trl}
Leandro von Werra, Younes Belkada, Lewis Tunstall, Edward Beeching, Tristan
  Thrush, Nathan Lambert, Shengyi Huang, Kashif Rasul, and Quentin
  Gallou{\'e}dec.
\newblock {TRL}: {T}ransformers reinforcement learning.
\newblock \url{https://github.com/huggingface/trl}, 2020.

\bibitem[Wainwright \& Jordan(2008)Wainwright and
  Jordan]{wainwright2008graphical}
Martin~J. Wainwright and Michael~I. Jordan.
\newblock Graphical models, exponential families, and variational inference.
\newblock \emph{Foundations and trends in machine learning}, 1\penalty0 (1--2),
  2008.

\bibitem[Wang et~al.(2025)Wang, Yu, Gao, Zheng, Liu, Lu, Dang, Chen, Yang,
  Zhang, Liu, Yang, Zhao, Yue, Song, Yu, Huang, and Lin]{wang_beyond_2025}
Shenzhi Wang, Le~Yu, Chang Gao, Chujie Zheng, Shixuan Liu, Rui Lu, Kai Dang,
  Xionghui Chen, Jianxin Yang, Zhenru Zhang, Yuqiong Liu, An~Yang, Andrew Zhao,
  Yang Yue, Shiji Song, Bowen Yu, Gao Huang, and Junyang Lin.
\newblock Beyond the 80/20 rule: high-entropy minority tokens drive effective
  reinforcement learning for {LLM} reasoning.
\newblock In \emph{Advances in neural information processing systems}, 2025.

\bibitem[West \& Potts(2025)West and Potts]{west_base_2025}
Peter West and Christopher Potts.
\newblock Base models beat aligned models at randomness and creativity.
\newblock In \emph{Conference on language modeling}, 2025.

\bibitem[Williams(1992)]{williams1992simple}
Ronald~J. Williams.
\newblock Simple statistical gradient-following algorithms for connectionist
  reinforcement learning.
\newblock \emph{Machine learning}, 8\penalty0 (3--4), 1992.

\bibitem[Xiao et~al.(2026)Xiao, Xia, Yang, Gao, Shen, Zhang, He, Lou, Luo,
  Wang, et~al.]{xiao2026mimo}
Bangjun Xiao, Bingquan Xia, Bo~Yang, Bofei Gao, Bowen Shen, Chen Zhang,
  Chenhong He, Chiheng Lou, Fuli Luo, Gang Wang, et~al.
\newblock {MiMo-V2-Flash} technical report.
\newblock \emph{arXiv preprint arXiv:2601.02780}, 2026.

\bibitem[Xing et~al.(2026)Xing, Wang, Gao, Li, and Tang]{xing2026trust}
Xingrun Xing, Haoqing Wang, Boyan Gao, Ziheng Li, and Yehui Tang.
\newblock Trust region on-policy distillation.
\newblock \emph{arXiv preprint arXiv:2606.01249}, 2026.

\bibitem[Yang et~al.(2025)Yang, Li, Yang, Zhang, Hui, Zheng, Yu, Gao, Huang,
  Lv, et~al.]{yang2025qwen3}
An~Yang, Anfeng Li, Baosong Yang, Beichen Zhang, Binyuan Hui, Bo~Zheng, Bowen
  Yu, Chang Gao, Chengen Huang, Chenxu Lv, et~al.
\newblock {Qwen3} technical report.
\newblock \emph{arXiv preprint arXiv:2505.09388}, 2025.

\bibitem[Yang et~al.(2026)Yang, Liu, Chen, Dai, Wang, Lin, Lee, Chen, Jiang,
  He, et~al.]{yang2026nemotron}
Zhuolin Yang, Zihan Liu, Yang Chen, Wenliang Dai, Boxin Wang, Sheng-Chieh Lin,
  Chankyu Lee, Yangyi Chen, Dongfu Jiang, Jiafan He, et~al.
\newblock {Nemotron-Cascade 2}: Post-training {LLM}s with cascade {RL} and
  multi-domain on-policy distillation.
\newblock \emph{arXiv preprint arXiv:2603.19220}, 2026.

\bibitem[Zhang et~al.(2026)Zhang, Yu, Chong, Sicilia, Tomz, Manning, and
  Shi]{zhang2025verbalized}
Jiayi Zhang, Simon Yu, Derek Chong, Anthony Sicilia, Michael~R. Tomz,
  Christopher~D. Manning, and Weiyan Shi.
\newblock Verbalized sampling: how to mitigate mode collapse and unlock {LLM}
  diversity.
\newblock In \emph{International conference on machine learning}, 2026.

\bibitem[Zhao et~al.(2026{\natexlab{a}})Zhao, Xin, Fan, Tong, Li, and
  Shen]{zhao2026decoupling}
Anhao Zhao, Haoran Xin, Yingqi Fan, Junlong Tong, Wenjie Li, and Xiaoyu Shen.
\newblock Decoupling {KL} and trajectories: a unified perspective for {SFT},
  {DAgger}, offline {RL}, and {OPD} in {LLM} distillation.
\newblock \emph{arXiv preprint arXiv:2605.16826}, 2026{\natexlab{a}}.

\bibitem[Zhao et~al.(2026{\natexlab{b}})Zhao, Xie, Liu, Huang, Pang, Chen, and
  Grover]{zhao_self-distilled_2026}
Siyan Zhao, Zhihui Xie, Mengchen Liu, Jing Huang, Guan Pang, Feiyu Chen, and
  Aditya Grover.
\newblock Self-distilled reasoner: on-policy self-distillation for large
  language models.
\newblock In \emph{International conference on machine learning},
  2026{\natexlab{b}}.

\bibitem[Zucchet et~al.(2025)Zucchet, Bornschein, Chan, Lampinen, Pascanu, and
  De]{zucchet2025facts}
Nicolas Zucchet, J{\"o}rg Bornschein, Stephanie Chan, Andrew Lampinen, Razvan
  Pascanu, and Soham De.
\newblock How do language models learn facts? {D}ynamics, curricula and
  hallucinations.
\newblock In \emph{Conference on language modeling}, 2025.

\bibitem[Zucchet et~al.(2026)Zucchet, Lee, and Linderman]{zucchet2026curse}
Nicolas Zucchet, Hyun~Dong Lee, and Scott Linderman.
\newblock Language models suffer from a curse of ambiguity.
\newblock In \emph{Advances in neural information processing systems}, 2026.

\end{thebibliography}
}
\newpage
\appendix

\renewcommand{\ptctitle}{Table of contents}
\setcounter{tocdepth}{3}
\setcounter{parttocdepth}{3}
\renewcommand \thepart{}
\renewcommand \partname{}
\renewcommand \thepart{Appendix}
\addcontentsline{toc}{section}{Appendix}
\part{}
\parttoc

\newpage

\section*{Notation}
\addcontentsline{toc}{section}{Notation}

\begin{center}
\renewcommand{\arraystretch}{1.15}
\setlength{\tabcolsep}{6pt}
\begin{tabular}{rl}
    \specialrule{0.12em}{0pt}{4pt}
    \textbf{Symbol} & \textbf{Meaning} \\
    \specialrule{0.12em}{3pt}{2pt}
    \multicolumn{2}{l}{\emph{Distillation framework (Section~\ref{sec:distillation})}} \\
    \midrule
    $x$ & prompt, sampled from a dataset \\
    $y$, $y_t$, $T$ & completion of the prompt, its $t$-th token and its length \\
    $y_{<t}$ & prefix of the completion before position $t$ \\
    $i = (x, y_{<t})$ & context used to compare the student and the teacher \\
    $d$ & vocabulary size \\
    $\pi_\theta(\cdot \mid i)$ & student distribution over the next token \\
    $\pi^*(\cdot \mid i)$ & teacher distribution over the next token \\
    $p(\cdot \mid x)$ & data-generating process the completions are sampled from \\
    $\ell$ & token-level objective, a divergence between next-token distributions \\
    $\mathcal{L}(\theta)$ & total loss, $\ell$ summed over tokens and averaged over completions \\
    $z(i) \in \mathbb{R}^d$ & logits of the student in context $i$ \\
    $\mathbb{E}_i$ & average over contexts, one per token \\
    $H(\pi)$ & Shannon entropy of a next-token distribution \\
    $\bar H(\pi)$ & average entropy over contexts, per token \\
    $D_\mathrm{KL}(a \parallel b)$ & Kullback--Leibler divergence, forward in $a$ and reverse in $b$ \\
    $D_f(a \parallel b)$ & $f$-divergence of $a$ from the reference distribution $b$ \\
    $g_\mathrm{seq}$ & prefix-visitation gradient the stop-gradient drops \\
    \midrule
    \multicolumn{2}{l}{\emph{Divergences and entropy (Section~\ref{sec:entropy})}} \\
    \midrule
    $\lambda$ & interpolation weight of the generalized Jensen--Shannon family \\
    $\mathrm{JS}_\lambda$ & generalized Jensen--Shannon divergence \\
    $m_\lambda = \lambda \pi^* + (1 - \lambda) \pi_\theta$ & mixture that the family compares both models to \\
    $\theta^*_\mathrm{for}$, $\theta^*_\mathrm{rev}$ & local optima of the forward and reverse KL objectives \\
    $\pi_{\theta^*_\mathrm{for}}$, $\pi_{\theta^*_\mathrm{rev}}$ & students at these local optima \\
    $k$ & number of tokens kept by a top-$k$ restriction \\
    $g(i)$ & mean-logit mismatch between the student and the teacher \\
    $\Delta(i)$ & excess surprisal of the student's samples under the teacher \\
    \midrule
    \multicolumn{2}{l}{\emph{Toy model (Section~\ref{subsec:toy_model})}} \\
    \midrule
    $h$ & embedding dimension \\
    $n$ & number of contexts \\
    $e_i \in \mathbb{R}^h$ & random embedding of context $i$ \\
    $W \in \mathbb{R}^{d \times h}$ & weight matrix of the student's linear head \\
    $\exp(H(\pi^*))$ & perplexity of the teacher, its effective support size \\
    \midrule
    \multicolumn{2}{l}{\emph{Diffuse high-load limit (Appendix~\ref{app:large-load-uniform})}} \\
    \midrule
    $E \in \mathbb{R}^{n \times h}$ & embedding matrix, stacking the $e_i^\top$ \\
    $P_E$ & orthogonal projector onto the column space of $E$ \\
    $u$ & uniform distribution over the vocabulary \\
    $\beta$ & inverse teacher temperature, the teacher being diffuse as $\beta \to 0$ \\
    $z^*(i)$, $z_\theta(i)$ & centered teacher and student logits in context $i$, stacked into $Z^*$, $Z_\theta$ \\
    \bottomrule
\end{tabular}
\end{center}

\newpage
\section{The distillation framework}
\label{app:derivations}

Section~\ref{sec:distillation} introduces the token-level distillation objective of Equation~\ref{eqn:objective} and states that writing the objective at the token level is what decouples the data-generating process $p$ from the divergence $\ell$.
This appendix makes that claim precise.
We first define the sequence-level objective, the token-level one and the surrogate that implementations actually differentiate, and compare them (Appendix~\ref{app:token-seq}).
We then discuss how the token-level objective is evaluated in practice (Appendix~\ref{app:full-vs-sampled}) and define the divergence families we sweep over (Appendix~\ref{app:families}).

\subsection{Token- and sequence-level objectives and where gradients are stopped}
\label{app:token-seq}

There are a few technical details in how distillation methods are defined, mostly when the completions are sampled from the student, that have important consequences.
Here, we emphasize the difference between token- and sequence-level objectives, and review how an implementation detail, the stop-gradient, alters the objective that is actually optimized.

\subsubsection{The different objectives}

\begin{definition}[Sequence-level objective]
\label{def:seq}
The sequence-level objective compares the student and teacher distributions over entire sequences,
\begin{equation}
    \mathcal{L}_\mathrm{seq}(\theta)=\mathbb{E}_x\left[\ell_\mathrm{seq}\left(\pi_\theta(\cdot\mid x),\pi^*(\cdot\mid x)\right)\right],
\end{equation}
where $\ell_\mathrm{seq}$ is a divergence between the two probability distributions over the full sequence space.
For the reverse KL divergence, $\ell_\mathrm{seq}=D_\mathrm{KL}(\pi_\theta(\cdot\mid x)\parallel \pi^*(\cdot\mid x))$, which hides an expectation over student-generated sequences.
The forward KL divergence instead hides an expectation over teacher-generated sequences.
\end{definition}

\begin{definition}[Token-level objective]
\label{def:tok}
The token-level objective is that of Equation~\ref{eqn:objective},
\begin{equation}
    \mathcal{L}_\mathrm{tok}(\theta)=\mathbb{E}_x\mathbb{E}_{y\sim p(\cdot\mid x)}\left[\sum_{t=1}^T\ell_\mathrm{tok}\left(\pi_\theta(\cdot\mid x,y_{<t}),\pi^*(\cdot\mid x,y_{<t})\right)\right],
\end{equation}
with $\ell_\mathrm{tok}$ a divergence between the token-level distributions in context $(x,y_{<t})$ and $p$ the data-generating process from which the sequences are sampled.
\end{definition}

\begin{definition}[Surrogate token-level objective]
\label{def:surtok}
The surrogate token-level objective ignores the potential dependency of the data-generating process on the parameters:
\begin{equation}
    \mathcal{L}_\mathrm{sur\text{-}tok}(\theta)=\mathbb{E}_x\mathbb{E}_{y\sim\text{stop-gradient}[p](\cdot\mid x)}\left[\sum_{t=1}^T\ell_\mathrm{tok}\left(\pi_\theta(\cdot\mid x,y_{<t}),\pi^*(\cdot\mid x,y_{<t})\right)\right].
\end{equation}
Its gradient is
\begin{equation}
    \nabla_\theta\mathcal{L}_\mathrm{sur\text{-}tok}=\mathbb{E}_x\mathbb{E}_{y\sim p}\left [\sum_t\nabla_\theta\ell_\mathrm{tok}(\pi_\theta(\cdot\mid x,y_{<t}),\pi^*(\cdot\mid x,y_{<t})) \right],
\end{equation}
and this is what most papers use in practice.
\end{definition}

\begin{remark}
\label{rem:surtok}
When $p=\pi^*$, the data-generating process does not depend on $\theta$ so the stop-gradient has no effect and $\mathcal{L}_\mathrm{sur\text{-}tok}$ and $\mathcal{L}_\mathrm{tok}$ have the same gradients.
The surrogate differs from the token-level objective only when sampling on-policy, i.e. $p=\pi_\theta$, or more generally when the sampling distribution depends on the student.
\end{remark}

\subsubsection{How the different objectives and their gradients compare}

\begin{fact}[The token- and sequence-level objectives can be equal]
\label{fact:tok-seq}
Take $p=\pi_\theta$ with $\ell_\mathrm{tok}$ and $\ell_\mathrm{seq}$ the reverse KL divergence, or $p=\pi^*$ with $\ell_\mathrm{tok}$ and $\ell_\mathrm{seq}$ the forward KL divergence.
Then $\mathcal{L}_\mathrm{tok}=\mathcal{L}_\mathrm{seq}$, and in particular $\nabla_\theta\mathcal{L}_\mathrm{tok}=\nabla_\theta\mathcal{L}_\mathrm{seq}$.
\end{fact}
\begin{proof}
The proof relies on the chain rule of the KL divergence, which we state for general autoregressive distributions before applying it to the setting of interest.
Let $a(\cdot\mid x)$ and $b(\cdot\mid x)$ be two distributions.
Their sequence-level log-ratio decomposes token by token,
\begin{equation}
    \log\frac{a(y\mid x)}{b(y\mid x)}=\sum_{t=1}^T\log\frac{a(y_t\mid x,y_{<t})}{b(y_t\mid x,y_{<t})}.
\end{equation}
Taking the expectation under $y\sim a(\cdot\mid x)$ and exchanging it with the finite sum,
\begin{equation}
    D_\mathrm{KL}(a(\cdot\mid x)\parallel b(\cdot\mid x))=\sum_{t=1}^T\mathbb{E}_{y\sim a(\cdot\mid x)}\left[\log\frac{a(y_t\mid x,y_{<t})}{b(y_t\mid x,y_{<t})}\right].
\end{equation}
The $t$-th summand depends on $y$ only through the prefix $y_{<t}$ and the current token $y_t$, so the future tokens $y_{>t}$ marginalize out (the tower property) and the expectation becomes
\begin{align}
    \mathbb{E}_{y\sim a(\cdot\mid x)}\left[\log\frac{a(y_t\mid x,y_{<t})}{b(y_t\mid x,y_{<t})}\right]
    &=\mathbb{E}_{y_{<t}\sim a(\cdot\mid x)} \, \mathbb{E}_{y_t\sim a(\cdot\mid x,y_{<t})}\left[\log\frac{a(y_t\mid x,y_{<t})}{b(y_t\mid x,y_{<t})}\right]\\
    &=\mathbb{E}_{y_{<t}\sim a(\cdot\mid x)}\left[D_\mathrm{KL}\left(a(\cdot\mid x,y_{<t})\parallel b(\cdot\mid x,y_{<t})\right)\right],
\end{align}
the inner expectation being the KL divergence between the token-level distributions in context $(x,y_{<t})$.
Summing over $t$ and collecting the per-step expectations back into a single expectation over $y\sim a(\cdot\mid x)$,
\begin{equation}
    D_\mathrm{KL}(a(\cdot\mid x)\parallel b(\cdot\mid x))=\mathbb{E}_{y\sim a(\cdot\mid x)}\left[\sum_{t=1}^T D_\mathrm{KL}\left(a(\cdot\mid x,y_{<t})\parallel b(\cdot\mid x,y_{<t})\right)\right].
\end{equation}
This has exactly the same form as the token-level objective of Definition~\ref{def:tok} with the data sampled from $a$ and $\ell_\mathrm{tok}$ the per-token KL divergence between $a$ and $b$.
The two subcases follow by choosing $(a,b)$.
\begin{itemize}
    \item[--] Taking $a=\pi_\theta$ and $b=\pi^*$ gives the reverse KL divergence with $p=a=\pi_\theta$.
    \item[--] Taking $a=\pi^*$ and $b=\pi_\theta$ gives the forward KL divergence with $p=a=\pi^*$.
\end{itemize}
In both, the outer expectation samples the prefixes from the first argument of the divergence, which is precisely $p$.
Equality of the objectives implies equality of their gradients, as there are no stop-gradients or additional implicit parameters held constant here.
\end{proof}

\begin{remark}
\label{rem:coupling}
Because the chain rule samples the prefixes from the first argument of the divergence, deriving an objective at the sequence level fixes the data-generating process at the token level: the reverse KL divergence forces $p=\pi_\theta$ and the forward KL divergence forces $p=\pi^*$.
This is why we argued that sequence-level objectives couple $\ell_\mathrm{tok}$ to $p$ in the main text.
Conversely, nothing couples the two once the objective is written at the token level from the start, which is what lets us sweep the divergence at a fixed data-generating process in Section~\ref{sec:entropy}.
\end{remark}

\subsubsection{What the stop-gradients used in practice ignore}

In practice, the surrogate objective is typically used.
It follows that the resulting gradient is a biased estimate of that of the actual objective.

\begin{fact}[The token-level gradient and its surrogate]
\label{fact:grad}
Write $\ell_t(y, \theta)=\ell_\mathrm{tok}(\pi_\theta(\cdot\mid x,y_{<t}),\pi^*(\cdot\mid x,y_{<t}))$ for the per-token loss along a sequence.
The token-level gradient decomposes as
\begin{equation}
    \nabla_\theta\mathcal{L}_\mathrm{tok}=\nabla_\theta\mathcal{L}_\mathrm{sur\text{-}tok}+g_\mathrm{seq},
\end{equation}
with the visitation gradient in its causal form
\begin{equation}
    g_\mathrm{seq}=\mathbb{E}_x\mathbb{E}_{y\sim p}\left[\sum_{s=1}^T\nabla_\theta\log p(y_s\mid x,y_{<s})\sum_{t=s}^T \ell_t(y, \theta)\right],
\end{equation}
each token weighted by its cost-to-go $\sum_{t=s}^T\ell_t$.
When $p=\pi^*$, $g_\mathrm{seq}=0$ and the three gradients coincide.
When $p=\pi_\theta$ it is generally nonzero, and only then does the surrogate the algorithm optimizes depart from the true gradient of the objective.
\end{fact}
\begin{proof}
Write $L(y,\theta)=\sum_t \ell_t(y, \theta)$ so that $\mathcal{L}_\mathrm{tok}=\mathbb{E}_x\mathbb{E}_{y\sim p}[L]$, and differentiate the sampled expectation with the score-function identity,
\begin{align}
    \nabla_\theta\mathbb{E}_{y\sim p}[L]&=\mathbb{E}_{y\sim p}\left[\nabla_\theta L\right]+\mathbb{E}_{y\sim p}\left[L\nabla_\theta\log p(y\mid x)\right],
\end{align}
using $\nabla_\theta p=p\,\nabla_\theta\log p$.
The first term differentiates only the losses, at fixed sampling distribution, and is $\nabla_\theta\mathcal{L}_\mathrm{sur\text{-}tok}$ of Definition~\ref{def:surtok}; the second is the visitation gradient.
When $p=\pi^*$ does not depend on $\theta$, this score term is absent and $g_\mathrm{seq}=0$.
It remains to bring the visitation gradient to the causal form of the statement.
The score factors over tokens,
\begin{equation}
    \nabla_\theta\log p(y\mid x)=\sum_{s=1}^T\nabla_\theta\log p(y_s\mid x,y_{<s}),
\end{equation}
and the losses $\ell_t$ with $t<s$ do not depend on the token drawn at step $s$ given its prefix, so, conditioning on $y_{<s}$ and using $\mathbb{E}_{y_s\sim p(\cdot\mid x,y_{<s})}[\nabla_\theta\log p(y_s\mid x,y_{<s})]=0$, their cross terms vanish in expectation.
Only the cost-to-go $\sum_{t=s}^T\ell_t$ survives against the step-$s$ score, giving the causal form
\begin{equation}
    g_\mathrm{seq}=\mathbb{E}_x\mathbb{E}_{y\sim p}\left[\sum_{s=1}^T\nabla_\theta\log p(y_s\mid x,y_{<s})\sum_{t=s}^T \ell_t(y,\theta)\right].
\end{equation}
\end{proof}

\begin{remark}
\label{rem:discount}
The per-token losses $\ell_t$ are per-step costs, and $g_\mathrm{seq}$ is the credit assigned to each token for its effect on the future costs it steers the sequence towards.
Keeping it in full is the undiscounted, full-horizon policy gradient, which corresponds to a discount factor of $\gamma=1$; the surrogate gradient drops it entirely, crediting each token only for its immediate cost, that is the discount factor is taken to be $\gamma=0$ \citep{thinkingmachines2025onpolicy}.
The myopic token-level algorithm used in practice is therefore the surrogate, not the sequence-level objective: the two objectives are equal (Fact~\ref{fact:tok-seq}), and it is only the gradient that is truncated.
This is why the surrogate dynamics do not in general follow the gradient of any well-defined objective \citep{czarnecki2019distilling}, the field $\nabla_\theta\mathcal{L}_\mathrm{sur\text{-}tok}$ being non-conservative once $p=\pi_\theta$ moves with $\theta$.
\end{remark}

\begin{remark}
\label{rem:variance}
Dropping $g_\mathrm{seq}$ is not only a modeling choice and can be a practical necessity.
As a score-function estimator over full rollouts, the gradient $g_\mathrm{seq}$ is high-variance and expensive to estimate, a difficulty at the heart of reinforcement learning with policy gradients \citep{williams1992simple, schulman2016gae}.
It is also the term that makes the student account for how its own choices reshape the sequence distribution on which it is later scored.
This is consistent with the reward hacking observed when solely optimizing the sequence-level reverse KL divergence \citep{gu2024minillm}.
\end{remark}

\begin{remark}
\label{rem:entropy-consequence}
This is the observation that lets the rest of the paper ignore the sequence-modeling aspects of distillation in our toy model of Section~\ref{subsec:toy_model}.
Once $g_\mathrm{seq}$ is dropped, every gradient the algorithm takes is an average of per-context gradients of $\ell_\mathrm{tok}$, and the contexts only enter through the distribution $p$ that generates them.
The toy model of Section~\ref{subsec:toy_model} keeps exactly this structure, a set of contexts and a per-context divergence, and discards the autoregressive coupling between them.
\end{remark}

\subsection{Practical evaluation of the token-level divergence}
\label{app:full-vs-sampled}

Evaluating the token-level divergence $\ell$ in Equation~\ref{eqn:objective} over the full vocabulary can be expensive.
Two strategies reduce this cost: estimating the divergence from a single sampled token, or restricting the sum to a top-$k$ subset of the vocabulary.

\subsubsection{Sampled-token evaluation}

Instead of summing over the vocabulary, one can estimate $\ell_t$ from a single sampled token, reducing the cost to a single teacher probability per context.
This is the only option when the teacher is implicit, as in next-token cross-entropy during pretraining, but is also used with explicit teachers for efficiency.
The gradient of this estimate needs care when the sample comes from the student.
We illustrate with the reverse KL, where $y_t \sim \pi_\theta$; the same applies to the student branch of the generalized Jensen--Shannon divergence but not to the forward KL, which samples from the teacher.

Naively differentiating the sampled log-ratio at fixed $y_t$ gives
\begin{equation}
    \nabla_\theta\log \frac{\pi_\theta(y_t\mid x,y_{<t})}{\pi^*(y_t\mid x,y_{<t})}=\nabla_\theta\log\pi_\theta(y_t\mid x,y_{<t}),
\end{equation}
since the teacher does not depend on $\theta$, and its expected value vanishes as
\begin{equation}
    \mathbb{E}_{y_t\sim\pi_\theta}[\nabla_\theta\log\pi_\theta(y_t\mid x,y_{<t})]=0,
    \label{eq:vocab-score}
\end{equation}
whereas $\nabla_\theta\ell_t$ is generally nonzero, because this estimate ignores that the distribution from which $y_t$ is drawn itself depends on $\theta$.
Accounting for it through the score-function identity
\begin{equation}
    \nabla_\theta\mathbb{E}_{y_t\sim\pi_\theta}[f]=\mathbb{E}_{y_t\sim\pi_\theta}[f\nabla_\theta\log\pi_\theta(y_t\mid x,y_{<t})+\nabla_\theta f]
\end{equation}
with $f$ the log-ratio, the integrand term $\mathbb{E}_{y_t\sim\pi_\theta}[\nabla_\theta f]$ is exactly the vanishing naive estimate above, so only the score-function term remains,
\begin{equation}
    \nabla_\theta\ell_t=\mathbb{E}_{y_t\sim\pi_\theta}\left[\log\frac{\pi_\theta(y_t\mid x,y_{<t})}{\pi^*(y_t\mid x,y_{<t})}\nabla_\theta\log\pi_\theta(y_t\mid x,y_{<t})\right].
    \label{eq:sampled-local-gradient}
\end{equation}
This is a second score function, taken at the output token level instead of the sequence level as in the prefix-visitation gradient $g_\mathrm{seq}$ of Fact~\ref{fact:grad}.

\paragraph{Practical implementations.}
The on-policy distillation recipe of \citet{thinkingmachines2025onpolicy} combines this approximation with the one of Remark~\ref{rem:discount}.
It estimates $\ell_t$ from the sampled token, using $-\operatorname{stop-grad}[\log(\pi_\theta(y_t\mid x,y_{<t})/\pi^*(y_t\mid x,y_{<t}))]$ as a per-token advantage inside a standard policy-gradient pipeline, and it truncates the cost-to-go to the immediate term ($\gamma=0$).
The first is the Monte Carlo approximation of the local loss gradient over the action space discussed above.
The second is the truncation of the sequence-level gradient over time that Appendix~\ref{app:token-seq} analyzes.
The two are independent of each other.

\subsubsection{Top-\texorpdfstring{$k$}{k} restrictions: renormalization and tail bucket}
\label{app:topk-definitions}

When the teacher's full probability vector is available, it is common to restrict the divergence to a subset $S \subseteq \{1, \ldots, d\}$ of the vocabulary, typically the top-$k$ tokens of the teacher or the student, for compute and bandwidth efficiency.
Given such a subset, there are two standard ways to handle the probability mass outside $S$.

\paragraph{Renormalization.}
Both distributions are renormalized over $S$ and the divergence is computed between the restricted distributions:
\begin{equation}
    \ell_{\text{renorm}}(\pi_\theta, \pi^*; S) = \ell\!\left(\frac{\pi_\theta|_S}{M_\theta}, \frac{\pi^*|_S}{M^*}\right),
\end{equation}
where $M_\theta = \sum_{j \in S} \pi_\theta(j)$ and $M^* = \sum_{j \in S} \pi^*(j)$ are the total masses on $S$, and $\pi|_S$ denotes the restriction of $\pi$ to $S$.
This discards the tail entirely: the gradient with respect to logits outside $S$ is zero, and the student receives no signal about how much total mass to place on the top-$k$ set.

\paragraph{Tail bucket.}
All mass outside $S$ is lumped into a single extra category and the divergence is computed over $k + 1$ tokens:
\begin{equation}
    \ell_{\text{tail}}(\pi_\theta, \pi^*; S) = \ell\!\left(\tilde\pi_\theta, \tilde\pi^*\right), \quad \text{with} \quad \tilde\pi(j) = \begin{cases} \pi(j) & j \in S, \\ 1 - \sum_{j' \in S} \pi(j') & j = \text{tail}. \end{cases}
\end{equation}
This preserves the total-mass constraint: the student is trained to match the teacher's tail mass as a lump sum, which provides one additional degree of freedom compared to renormalization.

As Section~\ref{subsec:topk} and Appendix~\ref{app:topk-additional} show, the choice between the two has a direct effect on the entropy of the student.

\subsection{The divergence families we consider}
\label{app:families}

Forward and reverse KL are the two main focuses of our analysis, but the transition between them is itself informative, so we also train on a family that interpolates between the two.
We use the generalized Jensen--Shannon divergence, the objective family of \citet{agarwal2024onpolicy} also used in other works \citep[e.g.,][]{hubotter2026rlsd}, which compares both models to their mixture
\begin{equation}
    m_\lambda = \lambda \pi^* + (1 - \lambda) \pi_\theta
\end{equation}
through
\begin{equation}
    \label{eqn:jsd}
    \mathrm{JS}_\lambda(\pi^*, \pi_\theta) = \lambda D_\mathrm{KL}(\pi^* \parallel m_\lambda) + (1 - \lambda) D_\mathrm{KL}(\pi_\theta \parallel m_\lambda),
\end{equation}
with $\lambda \in [0, 1]$.
The two KL divergences are recovered in the limits, after dividing by the prefactor that vanishes there.

As $\lambda \to 0$ the mixture tends to $\pi_\theta$, so 
\begin{equation}
    D_\mathrm{KL}(\pi_\theta \parallel m_\lambda) = O(\lambda^2)
\end{equation}
and
\begin{equation}
    \frac{\mathrm{JS}_\lambda}{\lambda} \to D_\mathrm{KL}(\pi^* \parallel \pi_\theta),
\end{equation}
which is the forward KL.

As $\lambda \to 1$ the mixture tends to $\pi^*$ and symmetrically
\begin{equation}
    \frac{\mathrm{JS}_\lambda}{1 - \lambda} \to D_\mathrm{KL}(\pi_\theta \parallel \pi^*),
\end{equation}
the reverse KL.

Both endpoints of Equation~\ref{eqn:jsd} are identically zero, so only the interior of the family has a gradient; we take the divergence at $\lambda = 0$ and $\lambda = 1$ to be its normalized limits, that is the exact forward and reverse KL, which makes the whole interval trainable, matching the implementations of \citet{agarwal2024onpolicy} and \citet{vonwerra2020trl}.
This scale mismatch is not a problem to compare learning dynamics and their observables (like entropy), as Adam \citep{kingma_adam_2015} is almost invariant to the overall scale of the loss \citep{balles_dissecting_2018}.

\paragraph{$f$-divergences.}
Throughout the theory, we will sometimes use results that hold for a broad class of divergences, $f$-divergences, which includes the generalized Jensen--Shannon family and its two endpoints.
We introduce them here for completeness and make the connection to the generalized Jensen--Shannon family explicit.
Given a convex function $f$ with $f(1) = 0$, the $f$-divergence of a distribution $a$ from a reference distribution $b$ is
\begin{equation}
    \label{eqn:fdiv}
    D_f(a \parallel b) = \sum_j b(j) f \left ( \frac{a(j)}{b(j)} \right ).
\end{equation}
The reverse KL is recovered with $f(t) = t \log t$, the forward KL with $f(t) = - \log t$, and the generalized Jensen--Shannon divergence with
\begin{equation}
    \label{eqn:fjsd}
    f_\lambda(t) = - \lambda \log \big ( \lambda + (1 - \lambda) t \big ) + (1 - \lambda) t \log \frac{t}{\lambda + (1 - \lambda) t}.
\end{equation}
What will matter to us is that all members of this class that are twice differentiable at $1$ share the same local geometry around $a = b$, namely the Fisher metric scaled by the curvature $f''(1)$, which is the statement of Appendix~\ref{app:equivalence} and what makes the diffuse limit of Appendix~\ref{app:large-load-uniform} divergence-independent.
That curvature is $f''(1) = 1$ for both KL directions and $f_\lambda''(1) = \lambda (1 - \lambda)$ for the generalized Jensen--Shannon divergence, which vanishes at the two endpoints and is brought back to $1$ by the $1 / \lambda$ and $1 / (1 - \lambda)$ normalizations above.

\newpage
\section{Theory}

\subsection{Divergence choice does not matter around optimality}
\label{app:equivalence}

We now show the claim of Section~\ref{sec:realizable}: at a fixed data-generating process, the forward and reverse KL objectives have the same gradient to first order around $\pi_\theta = \pi^*$.
The statement is per context, so fix a context $i$ and write $\pi_\theta$ and $\pi^*$ for the two next-token distributions there.
The forward KL gradient is
\begin{equation}
    \nabla_\theta D_\mathrm{KL}(\pi^* \parallel \pi_\theta) = - \sum_j \pi^*(j) \nabla_\theta \log \pi_\theta(j),
\end{equation}
the teacher being independent of $\theta$.
For the reverse KL, differentiating the product and using $\nabla_\theta \pi_\theta = \pi_\theta \nabla_\theta \log \pi_\theta$ gives
\begin{align}
    \nabla_\theta D_\mathrm{KL}(\pi_\theta \parallel \pi^*) &= \sum_j \nabla_\theta \pi_\theta(j) \log \left ( \frac{\pi_\theta(j)}{\pi^*(j)} \right ) + \sum_j \pi_\theta(j) \nabla_\theta \log \pi_\theta(j) \\
    &= \sum_j \pi_\theta(j) \nabla_\theta \log \pi_\theta(j) \log \left ( \frac{\pi_\theta(j)}{\pi^*(j)} \right ),
\end{align}
the second term vanishing by the vocabulary-level score identity of Equation~\ref{eq:vocab-score}.
Around optimality the log ratio linearizes as
\begin{equation}
    \log \left ( \frac{\pi_\theta(j)}{\pi^*(j)} \right ) = -\log \left ( \frac{\pi^*(j)}{\pi_\theta(j)} \right ) \approx 1 - \frac{\pi^*(j)}{\pi_\theta(j)},
\end{equation}
and substituting it, the constant term vanishes by the score identity again, leaving
\begin{equation}
    \nabla_\theta D_\mathrm{KL}(\pi_\theta \parallel \pi^*) \approx - \sum_j \pi^*(j) \nabla_\theta \log \pi_\theta(j) = \nabla_\theta D_\mathrm{KL}(\pi^* \parallel \pi_\theta).
\end{equation}
Averaging the two sides over contexts sampled from any data-generating process $p$ preserves the equality, as $p$ only weights the contexts and weights them identically on both sides.
This is why the statement is independent of $p$: the surrogate gradients of Definition~\ref{def:surtok} for the forward and the reverse KL agree to first order whatever the data-generating process is, as long as it is held fixed.
Near optimality, the choice between the two divergences is therefore immaterial, and whatever difference remains between two distillation runs has to come from the data they are trained on.

This equivalence is well known: it is a classical result from information geometry that all suitably standardized $f$-divergences, of which forward and reverse KL are special cases, agree up to second order around $\pi_\theta = \pi^*$, with the shared local quadratic form given by the Fisher information metric \citep{amari2016information, polyanskiy2025information}.
The whole generalized Jensen--Shannon family of Appendix~\ref{app:families} is therefore also equivalent to its two endpoints in this regime.

Importantly, the expansion above requires the perturbation $\pi_\theta - \pi^*$ to be small relative to $\pi^*$ on every token, that is $|\log(\pi_\theta(j) / \pi^*(j))|$ uniformly small.
This is a much stronger requirement than a small average KL divergence, since the KL weights log ratios by probability mass and is therefore blind to large relative perturbations on low-probability tokens, which is precisely where the gradients of the two objectives differ.

\subsection{Where the entropy gap changes sign}
\label{app:jsd-heatmaps}

\begin{figure}[]
    \centering
    \includegraphics{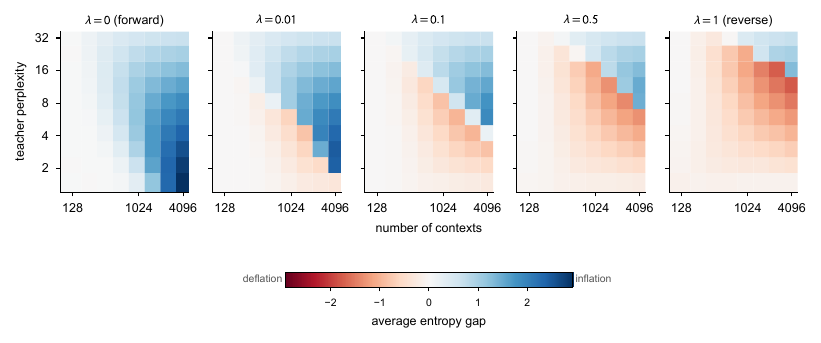}
    \vspace{-0.5cm}
    \caption{\textbf{The entropy gap phase transition moves with the divergence.}
    Average entropy gap $\bar H(\pi_\theta) - \bar H(\pi^*)$ at convergence over the plane of teacher perplexity and number of contexts in the setup of Section~\ref{sec:entropy}, at five weights $\lambda$ of the generalized Jensen--Shannon family, with the forward KL on the left and the reverse KL on the right.
    Each cell is its own configuration, averaged over $5$ seeds, and color indicates the entropy gap.
    The forward KL always inflates entropy and increasing $\lambda$ progressively moves the diagonal boundary, that is the task needs to be harder for the student to inflate entropy.}
    \label{fig:jsd_heatmaps}
\end{figure}

Figure~\ref{fig:toy_summary} varies the number of contexts, the teacher perplexity and the interpolation weight one at a time.
Figure~\ref{fig:jsd_heatmaps} repeats this experiment in the plane of number of contexts and teacher perplexity, over $5$ different generalized Jensen--Shannon divergence weights $\lambda$.
The results highlight that even though sweeping individual parameters yields sharp phase transitions, the boundary of the phase transition evolves much more smoothly.

\subsection{The forward KL inflates entropy}
\label{app:forward-entropy}

This appendix gives the formal backing for the claim in Section~\ref{sec:forward-kl} that the forward KL inflates entropy.
Assume the student family is closed under temperature scaling of the logits, which holds for any softmax model whose final linear layer can be rescaled, including standard language model heads.
As above we write $i$ for a context, $\pi_{\theta^*_\mathrm{for}}(\cdot \mid i)$ for the student at a stationary point $\theta^*_\mathrm{for}$ of the forward KL objective and $z(i)$ for its logits there, and take the contexts to be drawn from a data-generating process $p$ that is held fixed, as the surrogate objective of Definition~\ref{def:surtok} does.

Define $g(i) = \mathbb{E}_{\pi_{\theta^*_\mathrm{for}}}[z(i)] - \mathbb{E}_{\pi^*}[z(i)]$, the mean-logit mismatch in context $i$, which is also the derivative of the loss in $i$ when sharpening the logits along the temperature direction.
To see this, perturb the logits along the temperature direction, $z(i, \varepsilon) = (1 + \varepsilon) z(i)$, so that the loss in $i$ becomes
\begin{equation}
    \ell^\mathrm{for}(i, \varepsilon) = -(1+\varepsilon) \mathbb{E}_{\pi^*}[z(i)] + \log Z(i, \varepsilon)
\end{equation}
up to a term that does not depend on the student, with $Z(i, \varepsilon) = \sum_j \exp(z(i, \varepsilon)_j)$ the partition function of the perturbed student.
Using $\nabla_z \log Z = \pi_{\theta^*_\mathrm{for}}^{(\varepsilon)}$, the chain rule gives
\begin{equation}
    \frac{\mathrm{d}}{\mathrm{d}\varepsilon} \ell^\mathrm{for}(i, \varepsilon) = -\mathbb{E}_{\pi^*}[z(i)] + \mathbb{E}_{\pi_{\theta^*_\mathrm{for}}^{(\varepsilon)}}[z(i)],
\end{equation}
which at $\varepsilon = 0$ reduces to
\begin{equation}
    \frac{\mathrm{d}}{\mathrm{d}\varepsilon} \ell^\mathrm{for}(i, 0) = \mathbb{E}_{\pi_{\theta^*_\mathrm{for}}}[z(i)] - \mathbb{E}_{\pi^*}[z(i)] = g(i).
\end{equation}
At any interior stationary point of the objective, the derivative of the total loss along the temperature direction must vanish, which is exactly
\begin{equation}
    \label{eq:off-stationarity}
    \mathbb{E}_i[g(i)] = 0,
\end{equation}
the expectation being over contexts drawn from $p$, each token counting once as in Equation~\ref{eqn:objective}.

The difference in entropy between the student and the teacher can be rewritten to make its dependence on $g$ explicit:
\begin{align}
    H(\pi_{\theta^*_\mathrm{for}}) - H(\pi^*) &= - \mathbb{E}_{\pi_{\theta^*_\mathrm{for}}} \left [ \log \pi_{\theta^*_\mathrm{for}} \right ] + \mathbb{E}_{\pi^*} \left [ \log \pi^* \right ]\\
    &= \mathbb{E}_{\pi^*} \left [ \log \pi^* - \log \pi_{\theta^*_\mathrm{for}} \right ] +  \mathbb{E}_{\pi^*} \left [ \log \pi_{\theta^*_\mathrm{for}} \right ] - \mathbb{E}_{\pi_{\theta^*_\mathrm{for}}} \left [ \log \pi_{\theta^*_\mathrm{for}} \right ]\\
    &= D_\mathrm{KL}(\pi^* \parallel \pi_{\theta^*_\mathrm{for}}) + \mathbb{E}_{\pi^*} \left [ z(i) - \log Z(i) \right ] - \mathbb{E}_{\pi_{\theta^*_\mathrm{for}}} \left [ z(i) - \log Z(i) \right ]\\
    &= D_\mathrm{KL}(\pi^* \parallel \pi_{\theta^*_\mathrm{for}}) - g(i),
\end{align}
where the last equality comes from the fact that $\log Z$ is constant in the token and therefore cancels.
Leveraging the stationarity condition on $g$ (Equation~\ref{eq:off-stationarity}) yields
\begin{equation}
    \mathbb{E}_i \left [ H(\pi_{\theta^*_\mathrm{for}}) \right ] = \mathbb{E}_i \left [ H(\pi^*) \right ] + \mathbb{E}_i \left [ D_\mathrm{KL}(\pi^* \parallel \pi_{\theta^*_\mathrm{for}}) \right ].
\end{equation}
Note that nothing in the argument constrains $p$: the identity is a property of the forward KL alone, the data-generating process only setting which contexts the two averages run over.
When the teacher is implicit and only accessible through its samples, as in pretraining, the right-hand side is the cross-entropy loss, which gives the form of Equation~\ref{eqn:forward-kl-entropy} that Section~\ref{subsec:pretraining} tests.

This is an instance of a classical fact from information geometry: along its temperature direction, the student is a one-parameter exponential family with its own logits as sufficient statistic.
Stationarity of the forward KL divergence is moment matching of that statistic \citep{wainwright2008graphical}, and the entropy of a moment-matched projection exceeds the entropy of its target by exactly their divergence \citep{amari2016information}.

\paragraph{From average to pointwise inflation.}
So far, we have shown that entropy increases when averaging over contexts.
A natural follow-up question is whether the same holds in each context individually, that is, whether the entropy gap is always positive, or equivalently whether $g(i) < D_\mathrm{KL}(\pi^* \parallel \pi_{\theta^*_\mathrm{for}})$ everywhere.
The statement becomes pointwise as soon as the family contains per-context temperature directions, meaning that for every context the perturbation rescaling $z(i)$ while leaving the logits in other contexts unchanged is tangent to the family.
This is the case when the model has a learned input-dependent temperature on its logits, and approximately when the body is expressive enough to rescale its last-layer representation in one context without affecting the others.
Stationarity along these directions forces $g(i) = 0$ separately in every context, and the identity becomes
\begin{equation}
    H(\pi_{\theta^*_\mathrm{for}}) = H(\pi^*) + D_\mathrm{KL}(\pi^* \parallel \pi_{\theta^*_\mathrm{for}})
\end{equation}
in every context: the student entropy is strictly inflated wherever the student fails to match the teacher, by exactly the misfit there.
Note that per-context temperature scaling is far weaker than realizability, as a single scalar per context cannot match a full distribution, so the pointwise statement is not vacuous.
This average-versus-pointwise structure echoes the calibration literature, where stationarity along a single shared rescaling direction yields guarantees that hold on average over inputs \citep{clarte2023expectation}, while near-optimality against richer families of input-dependent post-processings yields correspondingly stronger pointwise ones \citep{blasiok2023when}.

\subsection{Reverse KL: a flatness statement}
\label{app:reverse-flatness}

A general statement on entropy similar to the one we obtained for the forward KL does not hold for the reverse one.
We apply the same recipe, decomposing the misfit and then exploiting stationarity along the temperature direction, but it yields only a flatness statement.
Writing $\pi_{\theta^*_\mathrm{rev}}$ for the student at a stationary point $\theta^*_\mathrm{rev}$ of the reverse KL objective and expanding the divergence as
\begin{equation}
    D_\mathrm{KL}(\pi_{\theta^*_\mathrm{rev}} \parallel \pi^*) = \mathbb{E}_{\pi_{\theta^*_\mathrm{rev}}}[\log \pi_{\theta^*_\mathrm{rev}}] - \mathbb{E}_{\pi_{\theta^*_\mathrm{rev}}}[\log \pi^*] = \mathbb{E}_{\pi_{\theta^*_\mathrm{rev}}}[-\log \pi^*] - H(\pi_{\theta^*_\mathrm{rev}}),
\end{equation}
adding $H(\pi_{\theta^*_\mathrm{rev}})$ and subtracting $H(\pi^*)$ and rearranging gives, in every context,
\begin{equation}
    \label{eq:on-decomposition}
    H(\pi_{\theta^*_\mathrm{rev}}) = H(\pi^*) + \Delta(i) - D_\mathrm{KL}(\pi_{\theta^*_\mathrm{rev}} \parallel \pi^*) \quad \text{with} \quad \Delta(i) = \mathbb{E}_{\pi_{\theta^*_\mathrm{rev}}}[-\log \pi^*] - H(\pi^*).
\end{equation}
The excess surprisal $\Delta(i)$ measures how much more surprising the student's samples are to the teacher than the teacher's own samples are.

The student is therefore less confident than the teacher exactly when $\Delta(i)$ exceeds the misfit $D_\mathrm{KL}(\pi_{\theta^*_\mathrm{rev}} \parallel \pi^*)$.
The temperature-direction argument no longer pins this sign down; it instead yields a flatness statement.
Perturbing the logits along the temperature direction, the derivative of the reverse KL loss in context $i$ is
\begin{equation}
    \frac{\mathrm{d}}{\mathrm{d}\varepsilon} \ell^\mathrm{rev}(i, 0) = \mathrm{Var}_{\pi_{\theta^*_\mathrm{rev}}}(\log \pi_{\theta^*_\mathrm{rev}}) - \mathrm{Cov}_{\pi_{\theta^*_\mathrm{rev}}}(\log \pi_{\theta^*_\mathrm{rev}}, \log \pi^*).
\end{equation}
Requiring the total derivative to vanish gives the stationarity condition on average over contexts,
\begin{equation}
    \label{eq:on-stationarity}
    \mathbb{E}_i \left [ \mathrm{Var}_{\pi_{\theta^*_\mathrm{rev}}}(\log \pi_{\theta^*_\mathrm{rev}}) \right ] = \mathbb{E}_i \left [
    \mathrm{Cov}_{\pi_{\theta^*_\mathrm{rev}}}(\log \pi_{\theta^*_\mathrm{rev}}, \log \pi^*) \right ].
\end{equation}
Bounding each covariance by Cauchy--Schwarz,
\begin{equation}
    \mathrm{Cov}_{\pi_{\theta^*_\mathrm{rev}}}(\log \pi_{\theta^*_\mathrm{rev}}, \log \pi^*) \leq
  \sqrt{\mathrm{Var}_{\pi_{\theta^*_\mathrm{rev}}}(\log \pi_{\theta^*_\mathrm{rev}}) \mathrm{Var}_{\pi_{\theta^*_\mathrm{rev}}}(\log \pi^*)},
\end{equation}
applying Cauchy--Schwarz a second time to the average over contexts, dividing by $\sqrt{\mathbb{E}_i [ \mathrm{Var}_{\pi_{\theta^*_\mathrm{rev}}}(\log \pi_{\theta^*_\mathrm{rev}}) ]}$ and squaring gives
\begin{equation}
      \mathbb{E}_i \left [ \mathrm{Var}_{\pi_{\theta^*_\mathrm{rev}}}(\log \pi_{\theta^*_\mathrm{rev}}) \right ] \leq \mathbb{E}_i \left [
  \mathrm{Var}_{\pi_{\theta^*_\mathrm{rev}}}(\log \pi^*) \right ].
\end{equation}
As a result, the same techniques that provided an entropy statement for the forward KL only provide a flatness statement here.

\subsection{Reverse KL inflates entropy in the diffuse high-load limit}
\label{app:large-load-uniform}

Empirically (Figure~\ref{fig:toy_summary}), we find that when the number of contexts is large (high load) or the teacher is diffuse (high perplexity), the reverse KL stops being mode-seeking and inflates entropy, as the forward KL does.
Here, we expand the distillation objective around a diffuse teacher and show that to leading order, forward KL, reverse KL and the whole generalized Jensen--Shannon family reduce to the same quadratic objective and therefore share the same solution.
This is why the top right corner of Figure~\ref{fig:jsd_heatmaps} in which both the teacher perplexity and the number of contexts are largest looks the same for every divergence.

\paragraph{The setup.}
We work in the toy model of Section~\ref{subsec:toy_model}, in which the teacher $\pi^*(\cdot \mid i)$ is drawn independently of the embedding $e_i$, from a law that is invariant under permutations of the vocabulary.
Stack the embeddings into $E \in \mathbb{R}^{n \times h}$, whose $i$-th row is $e_i^\top$, and write $P_E$ for the orthogonal projector onto its column space in $\mathbb{R}^n$.
For simplicity, we assume that $n > h$.
Random embeddings are in general position, so $\mathrm{rank}(E) = h$.
The student head produces the logit matrix $Z_\theta = E W^\top \in \mathbb{R}^{n \times d}$, so each of its columns is constrained to the $h$-dimensional column space of $E$.
The ratio $h / n$ is therefore the fraction of the $n$ context directions the student can exactly represent.

\paragraph{The diffuse limit.}
We make the teacher diffuse, keeping $h / n$ arbitrary.
We model it by fixing the teacher logits $z^*(i)$ and scaling them down by a factor $\beta \to 0$, so that the teacher distribution becomes uniform in the limit.
Write the teacher and the student in centered logit coordinates,
\begin{align}
    \pi^*(\cdot \mid i) &= \mathrm{softmax} \big ( \beta z^*(i) \big ), \\
    \pi_\theta(\cdot \mid i) &= \mathrm{softmax} \big ( z_\theta(i) \big ), \\
    \mathbf{1}^\top z^*(i) &= \mathbf{1}^\top z_\theta(i) = 0,
\end{align}
with the logits $z^*(i)$ fixed and drawn independently across contexts.

All $f$-divergences that are twice differentiable at $1$ agree to second order around coinciding distributions, and their common quadratic form is the Fisher metric scaled by $f''(1)$ \citep{amari2016information}; this is the argument we used in Appendix~\ref{app:equivalence} for the two KL directions.
In logit coordinates it reads
\begin{equation}
    \label{eq:fisher-logit}
    D_f \big ( \mathrm{softmax}(z + \delta) \parallel \mathrm{softmax}(z) \big ) = \frac{f''(1)}{2} \mathrm{Var}_{\mathrm{softmax}(z)}(\delta) + O(\lVert \delta \rVert^3),
\end{equation}
the variance being over tokens drawn from the reference distribution, the second argument of $D_f$.
The expansion holds as long as the offset $\delta$ is small, which here means that all the logits involved are $O(\beta)$: the teacher's by construction, and the student's because we expand around the branch on which it stays that close to uniform.
We apply it with the teacher as the reference and $\delta = z_\theta(i) - \beta z^*(i)$.
The teacher is itself $O(\beta)$ away from the uniform distribution, so $\mathrm{Var}_{\pi^*}$ can in turn be replaced by $\mathrm{Var}_u$ at the price of a further $O(\beta^3)$ term, and for centered logits $\mathrm{Var}_u(\delta) = \lVert \delta \rVert^2 / d$.
Stacking the logits into $Z^*, Z_\theta \in \mathbb{R}^{n \times d}$, this gives
\begin{equation}
    \label{eq:fdiv-large-load-taylor}
    \mathbb{E}_i \left [ D_f \big ( \pi_\theta(\cdot \mid i) \parallel \pi^*(\cdot \mid i) \big ) \right ] = \frac{f''(1)}{2 n d} \lVert Z_\theta - \beta Z^* \rVert_F^2 + O(\beta^3).
\end{equation}
Both KL directions are of this form, with $\pi^*$ as the reference distribution and $f''(1) = 1$, so they have the same minimizer.
The same argument applies to the generalized Jensen--Shannon family of Appendix~\ref{app:families}.

Minimizing Equation~\ref{eq:fdiv-large-load-taylor} over the achievable logits $Z_\theta = E W^\top$ is a least-squares projection onto the column space of $E$, so the optimal student logits are
\begin{equation}
    \label{eq:large-load-optimum}
    Z_\theta^\star = \beta P_E Z^* + O(\beta^2).
\end{equation}
The teacher's logits have scale $O(\beta)$, the fitted ones only $O(\beta \sqrt{h / n})$: the entries of $Z^*$ are independent and $P_E$ is a projector of rank $h$ that does not depend on them, so
\begin{equation}
    \mathbb{E}_{Z^*} \lVert P_E Z^* \rVert_F^2 = \frac{h}{n} \mathbb{E}_{Z^*} \lVert Z^* \rVert_F^2.
\end{equation}
As a result, increasing the number of contexts while keeping the model size fixed, which is what we do in our experiments, makes the student logits smaller relative to the teacher's, and the student distribution closer to uniform.

\paragraph{Entropy inflation.}
Applying Equation~\ref{eq:fisher-logit} again, this time with the uniform distribution as the reference and combining it with $D_\mathrm{KL}(\pi \parallel u) = \log d - H(\pi)$ provides an entropy expansion: a distribution $\mathrm{softmax}(\delta)$ with small centered logits has entropy 
\begin{equation}
    H(\mathrm{softmax}(\delta)) = \log d - \lVert \delta \rVert^2 / (2d) + O(\lVert \delta \rVert^3).
\end{equation}
Applying this to the teacher and to the student at the optimum of Equation~\ref{eq:large-load-optimum},
\begin{equation}
    \label{eq:diffuse-entropies}
    \bar H(\pi^*) = \log d - \frac{\beta^2}{2 n d} \lVert Z^* \rVert_F^2 + O(\beta^3), \qquad \bar H(\pi_\theta) = \log d - \frac{\beta^2}{2 n d} \lVert P_E Z^* \rVert_F^2 + O(\beta^3),
\end{equation}
so that
\begin{equation}
    \label{eq:large-load-entropy-overshoot}
    \bar H(\pi_\theta) - \bar H(\pi^*) = \frac{\beta^2}{2 n d} \lVert (I - P_E) Z^* \rVert_F^2 + O(\beta^3).
\end{equation}
That is, for small enough $\beta$ values, the student inflates entropy relative to the teacher.

If we further assume that the numbers of contexts and tokens are large enough that we can replace the teacher-related quantities by their expected values, the entropy gap only depends on the student through the single ratio $h / n$.
Indeed
\begin{equation}
    \mathbb{E}_{Z^*} \lVert (I - P_E) Z^* \rVert_F^2 = \left (1 - \frac{h}{n} \right ) \mathbb{E}_{Z^*} \lVert Z^* \rVert_F^2,
\end{equation}
so that Equation~\ref{eq:large-load-entropy-overshoot} becomes
\begin{align}
    \bar H(\pi_\theta) - \bar H(\pi^*) &= \frac{\beta^2}{2 n d} \left ( 1 - \frac{h}{n} \right ) \mathbb{E}_{Z^*} \lVert Z^* \rVert_F^2 + O(\beta^3) \nonumber \\
    &= \left ( 1 - \frac{h}{n} \right ) \left ( \log d - \bar H(\pi^*) \right ) + O(\beta^3),
    \label{eq:merged-entropy}
\end{align}
the second line recognizing in the first the teacher's entropy deficit below uniform of Equation~\ref{eq:diffuse-entropies}.
It follows that, in the diffuse high-load limit, the gap increases as the number of contexts $n$ increases, and decreases as the perplexity of the teacher increases, consistently with Figure~\ref{fig:toy_summary}b and~\ref{fig:toy_summary}c.

Evaluating Equation~\ref{eq:fdiv-large-load-taylor} at the optimum and comparing it to Equation~\ref{eq:large-load-entropy-overshoot} gives one last identity,
\begin{equation}
    \mathbb{E}_i \left [ D_f \big ( \pi_\theta(\cdot \mid i) \parallel \pi^*(\cdot \mid i) \big ) \right ] = f''(1) \left ( \bar H(\pi_\theta) - \bar H(\pi^*) \right ) + O(\beta^3),
\end{equation}
so for either KL direction the residual divergence equals the entropy inflation.
For the forward KL this is Equation~\ref{eqn:forward-kl-entropy}, which holds for any number of contexts and any teacher; reverse KL obeys it too in the diffuse limit.

\subsection{Additional results on top-\texorpdfstring{$k$}{k}}
\label{app:topk-additional}

\begin{figure}[]
    \centering
    \includegraphics{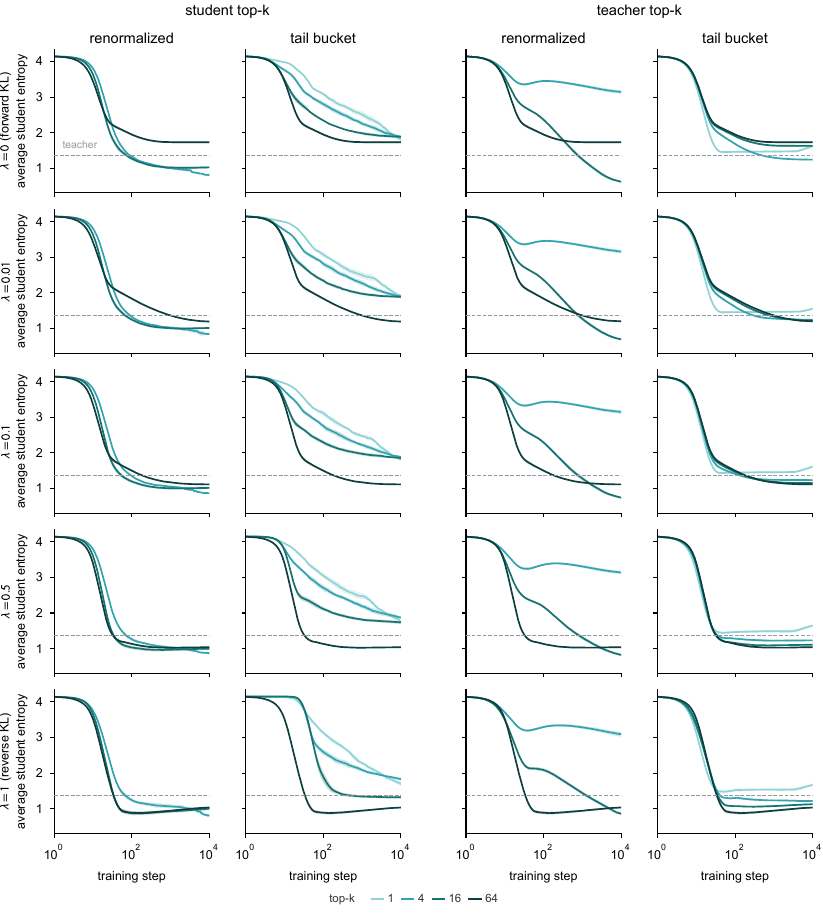}
    \caption{\textbf{The top-$k$ approximation across the divergence family.}
    Average student entropy over training, for the two choices the approximation makes: which model defines the support, i.e. whether we take the student's own top-$k$ or the teacher's, and what happens to the mass left outside the support, either renormalizing both sides over the window or keeping it in one extra tail bucket.
    Rows are the generalized Jensen--Shannon weight, from the forward KL at $\lambda = 0$ to the reverse KL at $\lambda = 1$.
    Curves are averaged over $5$ seeds $\pm \mathrm{std}$, and the dashed line is the teacher entropy.}
    \label{fig:topk_alpha_seeds}
\end{figure}

Figure~\ref{fig:topk_alpha_fkl} studies the effect of the different top-$k$ approximations for the forward KL only.
The forward KL is the $\lambda = 0$ endpoint of the generalized Jensen--Shannon family.
Figure~\ref{fig:topk_alpha_seeds} repeats the experiment at five different generalized Jensen--Shannon weights of Figure~\ref{fig:jsd_heatmaps}.
The vocabulary size is $64$, so $k = 64$ is the untruncated objective we use in the rest of the toy model experiments.
Renormalizing a top-$1$ distribution gives a Dirac on one token for both student and teacher, so the divergence is identically zero and no learning occurs; we therefore remove this configuration from all our experiments.

\subsection{Setup of the toy model experiments}
\label{app:toy-setup}

This section details the toy model of Section~\ref{subsec:toy_model} and the simulations of Figures~\ref{fig:toy_summary}, \ref{fig:topk_alpha_fkl}, \ref{fig:jsd_heatmaps} and~\ref{fig:topk_alpha_seeds}.

\paragraph{Model.}
The vocabulary has $d = 64$ tokens and the embeddings have dimension $h = 128$.
Each of the $n$ contexts is assigned an embedding $e_i$ drawn from a standard Gaussian distribution in $\mathbb{R}^h$ and normalized to unit norm, which stays fixed during training.
The student is a linear layer followed by a softmax, $\pi_\theta(\cdot \mid i) = \mathrm{softmax}(W e_i + b)$, with $W \in \mathbb{R}^{d \times h}$ initialized with LeCun normal initialization and $b$ initialized at zero.
The main text omits the bias $b$ for simplicity.

\paragraph{Teachers.}
For each context $i$, we draw a target perplexity $P_i$ and Gaussian logits $\zeta_i \sim \mathcal{N}(0, I_d)$, and set $\pi^*(\cdot \mid i) = \mathrm{softmax}(\zeta_i / \tau_i)$, with the temperature $\tau_i$ chosen so that $H(\pi^*(\cdot \mid i)) = \log P_i$.
By default, the perplexities are drawn independently and log-uniformly in $[1, 16]$, so that contexts differ both in which tokens the teacher favors and in how diffuse it is.
The teacher perplexity sweep of Figure~\ref{fig:toy_summary}c and the heatmaps of Figure~\ref{fig:jsd_heatmaps} instead give the same perplexity to all contexts.
All contexts have the same weight in the objective.

\paragraph{Training.}
Every step uses all contexts, and the student is trained with Adam \citep{kingma_adam_2015} and a constant learning rate of $0.1$ for $40{,}000$ steps.
All results are averaged over $5$ seeds, each of which redraws the embeddings, the teachers and the initialization.

\paragraph{Sweeps.}
The default number of contexts is $n = 2048$.
Figure~\ref{fig:toy_summary}b uses $21$ values of $n$ log-spaced between $192$ and $67{,}584$, Figure~\ref{fig:toy_summary}c uses $21$ perplexities log-spaced between $1$ and $32$, and Figure~\ref{fig:toy_summary}d uses $21$ weights $\lambda$ log-spaced between $0.001$ and $0.99$.
The trajectories of Figure~\ref{fig:toy_summary}e and the heatmaps of Figure~\ref{fig:jsd_heatmaps} use $\lambda \in \{0, 0.01, 0.1, 0.5, 1\}$, the heatmaps over $10$ perplexities log-spaced between $1.5$ and $32$ and $8$ numbers of contexts log-spaced between $128$ and $4096$.
The top-$k$ experiments of Figures~\ref{fig:topk_alpha_fkl} and~\ref{fig:topk_alpha_seeds} use $n = 512$ contexts and $k \in \{1, 4, 16, 64\}$, and Figure~\ref{fig:topk_alpha_seeds} uses the same five weights $\lambda$ as Figure~\ref{fig:jsd_heatmaps}.

\newpage
\section{Additional experimental results}
\label{app:additional}

\subsection{The identity across model families and pretraining stages}
\label{app:pretraining-identity}

\begin{figure}[h]
    \centering
    \includegraphics{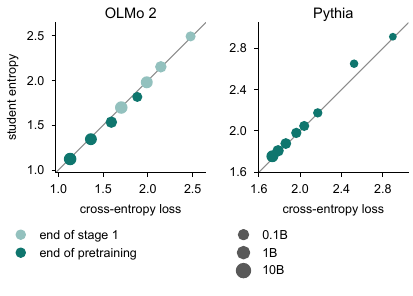}
    \caption{\textbf{At the end of pretraining, a model's mean entropy equals its mean cross-entropy.}
    Similar to Figure~\ref{fig:pretraining_sft} (left), for different model families and at different stages of pretraining.
    (left) OLMo~2 at 1B, 7B, 13B and 32B, at the end of stage~1 of pretraining (light) and at the end of pretraining (dark).
    (right) the eight Pythia models from 70M to 12B on the Pile \citep{gao2020pile}.}
    \label{fig:pretraining_pythia_olmo}
\end{figure}

Figure~\ref{fig:pretraining_sft} only covers one family of models, OLMo~2 \citep{olmo_2_2025}, and only at the end of its pretraining.
Figure~\ref{fig:pretraining_pythia_olmo} extends it to the Pythia family of models \citep{biderman_pythia_2023} and to an earlier point of OLMo~2's pretraining, averaging entropy and cross-entropy over about two million tokens per checkpoint.

OLMo~2 is pretrained in two stages, a long first one on a large web-dominated mixture and a short second one on a smaller and higher quality mixture, over which the learning rate is annealed to zero.
The light points are the checkpoints at the end of the first stage, measured on the first-stage mixture, and the dark ones the final pretrained models, measured on the second-stage mixture, which is the configuration Figure~\ref{fig:pretraining_sft} reports.
The pretraining mixture of Figure~\ref{fig:pretraining_sft} is therefore the second-stage mixture, the last data the pretrained models were trained on.
We draw the same number of tokens from every source of the mixture, but average entropy and cross-entropy at the published proportions of that mixture.

We use the standard Pythia models rather than their deduplicated variants.
They are trained in a single stage on the Pile \citep{gao2020pile}, whose training stream is released pre-tokenized and in the order the models consumed it, so the tokens we measure on are exactly the ones the models were trained on.
Since that run is a single epoch, position in the stream is position in training time, and we sample three chunks of equal size from its beginning, middle and near its end.

\subsection{Additional on- vs. off-policy distillation results}
\label{app:teacher-dist-results}

\begin{figure}[h]
    \centering
    \includegraphics{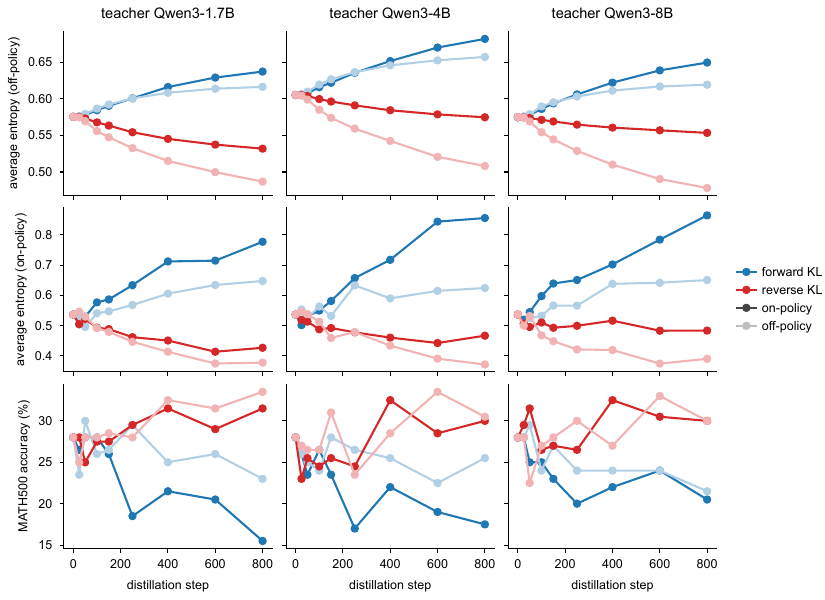}
    \caption{\textbf{Teacher-to-student distillation from non-thinking teachers.}
    Average entropy of the student on frozen teacher completions (top) and on its own completions (middle, as in Figure~\ref{fig:teacher-dist}), and MATH500 accuracy (bottom), against distillation step, for teachers of increasing size (left to right).
    Color is the divergence, with the full color for on-policy and a lighter shade for off-policy.
    Each curve is a single seed.}
    \label{fig:teacher-dist-nt}
\end{figure}

Figure~\ref{fig:teacher-dist} only reports the entropy of the student on its own completions.
Figure~\ref{fig:teacher-dist-nt} completes it with the entropy of the student on frozen teacher completions, where the contexts cannot move and only the weights can, and with its accuracy.
Accuracy is the in-training evaluation, averaged over one sample on $200$ MATH500 problems at the evaluation budget of Table~\ref{tab:teacher-dist-hyperparameters}, and is therefore noisier than the endpoint evaluation.

On frozen teacher completions, forward KL inflates entropy and reverse KL deflates it, and off-policy reverse KL deflates it most.
On the student's own completions the split has the same sign, with on-policy forward KL inflating entropy most.
Reverse KL slightly improves accuracy over the base student while forward KL, in particular on-policy, degrades it.

The lower accuracy of forward KL may largely be a consequence of the different levels of entropy it reaches rather than of what the student learns.
The in-training evaluation samples at temperature $1$, and a student with more entropy is more likely to sample a wrong token along a long solution.
Consistently, at the lower sampling temperature of the endpoint evaluation (Appendix~\ref{app:teacher-dist-setup}), all twelve runs score above the base student on MATH500, between $43.00\%$ and $46.27\%$ against $42.87\%$.
We expect accuracy to vary smoothly with entropy in this regime, so that the sampling temperature, or the other hyperparameters, could likely be adjusted to recover part of this gap.
We did not try to: we kept the hyperparameters of \citet{zhao2026decoupling} fixed across the grid and did not tune them for accuracy, as our goal is to study entropy.
These results should therefore only be read as showing that some of these settings improve the accuracy of the student, not as a comparison of the best accuracy each objective can reach.

\subsection{Setup of the on- vs. off-policy distillation experiments}
\label{app:teacher-dist-setup}

These experiments vary the sampling distribution and the divergence independently, as in Section~\ref{sec:distillation}, on a standard reasoning distillation task.
Following \citet{zhao2026decoupling}, we distill the Qwen3 instruct models with thinking disabled, at $1.7$B, $4$B and $8$B parameters, into Qwen3-0.6B-Base \citep{yang2025qwen3}, on the math problems of DeepScaleR \citep{luo2025deepscaler}.
The prefixes on which the token-level divergence is evaluated are either completions sampled from the teacher once and cached before training (off-policy), or completions sampled from the current student at every step (on-policy).
The divergence is either the forward or the reverse KL, computed on the full vocabulary at every position of the prefix.
Crossing the two prefix sources with the two divergences gives four objectives, and with three teachers the grid consists of twelve runs, all starting from the same base student.
Table~\ref{tab:teacher-dist-hyperparameters} lists the hyperparameters, which follow \citet{zhao2026decoupling} except for the choices motivated below.

\begin{table}[t]
    \centering
    \caption{\textbf{Hyperparameters of the on- vs. off-policy distillation experiments.}
    Values not motivated in the text follow \citet{zhao2026decoupling}, with three teachers rather than their two and one seed rather than their three.}
    \label{tab:teacher-dist-hyperparameters}
    \vspace{0.3cm}
    \begin{tabular}{@{}ll@{}}
        \toprule
        \multicolumn{2}{@{}l}{\textit{Setup}} \\
        \midrule
        Trainer & general online logit distillation trainer of TRL 1.9.2 \citep{vonwerra2020trl} \\
        Reference implementation & \citet{zhao2026decoupling} \\
        Generation engine & vLLM \citep{kwon2023vllm}, colocated \\
        Dataset & DeepScaleR \citep{luo2025deepscaler} \\
        Student model & Qwen3-0.6B-Base \citep{yang2025qwen3} \\
        Teacher models & Qwen3-1.7B, Qwen3-4B, Qwen3-8B, thinking disabled \\
        Prompt frame & problem text and boxed-answer instruction, no chat template \\
        Completions generated & $40000$ per teacher \\
        Usable targets & $25694$ to $27393$ \\
        \midrule
        \multicolumn{2}{@{}l}{\textit{Teacher corpus}} \\
        \midrule
        Sampling temperature & $1.0$ \\
        Top-$p$, top-$k$ & none \\
        Generation budget & $4096$ \\
        Truncation rule & first balanced boxed answer \\
        Completions without answer & dropped from training \\
        Filtering on correctness & none \\
        \midrule
        \multicolumn{2}{@{}l}{\textit{Objective}} \\
        \midrule
        Divergence & forward and reverse KL \\
        Prefix source & cached teacher corpus, or on-policy student rollouts \\
        Token restriction & none, full vocabulary \\
        Maximum prompt length & $1024$ \\
        Maximum completion length & $4096$ \\
        \midrule
        \multicolumn{2}{@{}l}{\textit{Optimization}} \\
        \midrule
        Optimizer & AdamW, weight decay $0$, clipping $1.0$ \\
        Learning rate & $5 \cdot 10^{-7}$ \\
        Schedule & constant, $45$ warmup steps \\
        Optimizer steps & $800$ \\
        Questions per optimizer step & $32$ \\
        Seeds & $1$ \\
        Precision & bfloat16 \\
        Hardware & 1 H100 \\
        \midrule
        \multicolumn{2}{@{}l}{\textit{In-training evaluation}} \\
        \midrule
        Entropy measured on & frozen teacher completions, and student rollouts \\
        Evaluation sets & held-out DeepScaleR, MATH500 \citep{lightman2024verify} \\
        Evaluation sampling & temperature $1.0$, top-$p$ $1.0$, pinned seed \\
        Evaluation budget & $2048$ tokens \\
        \midrule
        \multicolumn{2}{@{}l}{\textit{Endpoint evaluation}} \\
        \midrule
        Sampling & temperature $0.6$, top-$p$ $0.95$, $8192$ tokens \\
        Benchmarks & MATH500, GSM8K ($3$ samples), AMC23 ($5$ samples) \\
        Verifier & symbolic equivalence, with a numeric fallback \\
        \bottomrule
    \end{tabular}
\end{table}

\paragraph{The prompt frame.}
A token-level divergence compares the two models on the same prefix, so they have to share a prompt frame, yet they work best in different ones.
The base student degrades inside the chat template, from $42.87\%$ to $14.93\%$ accuracy on MATH500 \citep{lightman2024verify}, whereas the instruct teachers remain accurate outside of it but almost never stop, as there is no turn to end.
We therefore prompt both models without chat template, with the problem text followed by the instruction ``Let's think step by step and output the final answer within \texttt{\textbackslash boxed\{\}}.'', cut each teacher completion right after its first boxed answer, and append the end-of-text token of the student.
This avoids a format warmup stage, which would amount to off-policy forward KL distillation, that is, to one of the four objectives being compared.

\paragraph{The teacher corpus and the training length.}
Teacher completions are sampled at temperature $1$ without top-$p$ or top-$k$ truncation, unlike in \citet{zhao2026decoupling}, because the objective is an expectation under the distribution of the teacher.
Completions without a boxed answer within $4096$ tokens are dropped, which keeps about two thirds of them and biases the corpus towards problems that the teacher solves concisely.
Training runs for $800$ steps of $32$ prompts, just below one pass over the smallest corpus, so that no teacher benefits from more repetitions than another.
We add Qwen3-1.7B to the two teachers of \citet{zhao2026decoupling} because Qwen3-4B and Qwen3-8B are statistically indistinguishable on AMC23 ($66.2\%$ against $66.6\%$ accuracy).

\paragraph{Evaluation.}
We measure the entropy of the student on its own completions, whose contexts drift during training, and on a fixed set of cached teacher completions, on which only the weights of the student can change.
Evaluation completions are generated with a fixed seed, so that neighboring evaluation points share contexts.
We report the entropy pooled across tokens, as \citet{zhao2026decoupling} do.
As it weights each completion by its length, on the student's own completions it also moves when which completions are long changes, and the per-sequence average, which we also log, can disagree with it in sign.
The frozen MATH500 completions keep all $200$ problems, whereas the held-out DeepScaleR problems go through the same filtering as the training corpus and differ from one teacher to the other, so we compare teachers on MATH500 only.
Endpoint accuracy is measured at temperature $0.6$ and top-$p$ $0.95$ with a budget of $8192$ tokens, averaged over three samples per problem.

\paragraph{Pure bfloat16 training freezes the normalization gains.}
Following \citet{zhao2026decoupling}, we train in bfloat16 without float32 master weights.
All normalization gains then stay bit-identical throughout training, as the bfloat16 spacing at their typical magnitude is four orders of magnitude above the optimizer step, while the weight matrices do move.
As the final normalization gain controls the scale of the logits, these runs cannot test any statement about that scale, and we make none.

\subsection{Accuracy vs. entropy for self-distillation on the Chemistry dataset}

\begin{figure}[h]
    \centering
    \includegraphics{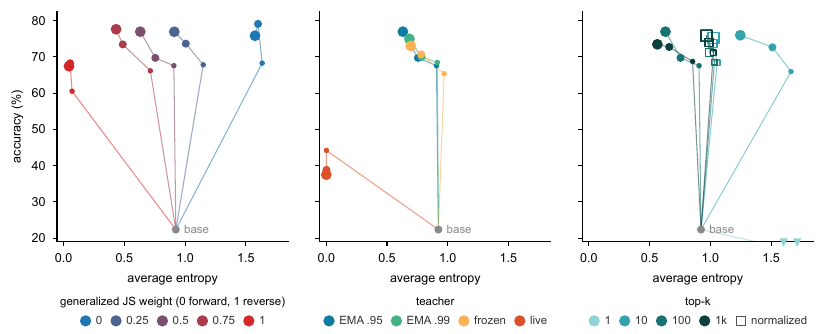}
    \caption{\textbf{Accuracy of the student model as a function of its entropy, in the self-distillation experiment of Figure~\ref{fig:self-dist}}.
    Each dot is a different checkpoint, the larger the dot, the later in training it is.
    JS stands for Jensen--Shannon and EMA for exponential moving average.
    }
    \label{fig:self-dist-accuracy}
\end{figure}

Figure~\ref{fig:self-dist} only reported the entropy of the student over time on the Chemistry dataset.
We report the corresponding accuracy in Figure~\ref{fig:self-dist-accuracy} for the sake of completeness.

\subsection{Additional self-distillation experiments on GSM8K}
\label{app:self-dist-gsm8k}

\begin{figure}[h]
    \centering
    \includegraphics{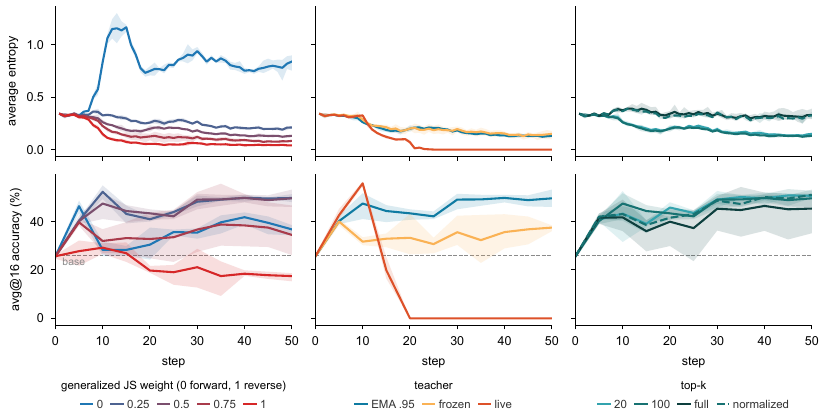}
    \caption{\textbf{Self-distillation on GSM8K behaves as on the Chemistry dataset.}
    Entropy of the student (top) and accuracy, averaged over $16$ samples per question (bottom), against training step, varying the weight of the generalized Jensen--Shannon (JS) divergence (left), the teacher (center), and how many tokens are taken into account and whether a tail bucket is used for the remaining tokens or not (right); defaults are $\lambda = 0.5$, an exponential moving average (EMA) teacher with decay $0.95$, and top-$100$ with tail bucket.
    Lines are means over three seeds and bands their range, and the dashed line is the accuracy of the base model.}
    \label{fig:self-dist-gsm}
\end{figure}

The main text follows the experimental setup of \citet{hubotter2026rlsd} on the Chemistry dataset.
We repeat it here on GSM8K \citep{cobbe2021gsm8k} with Qwen3-0.6B \citep{yang2025qwen3}, a testbed cheap enough to sweep the learning rate and to train ten times longer; Table~\ref{tab:self-dist-hyperparameters} lists the hyperparameters of both setups.
Figure~\ref{fig:self-dist-gsm} is the counterpart of Figure~\ref{fig:self-dist}, at a learning rate of $10^{-5}$, and shows the same qualitative behavior: entropy decreases monotonically with the generalized Jensen--Shannon weight $\lambda$, a teacher that follows the student too closely collapses, and restricting the divergence to fewer tokens raises entropy.

\begin{figure}[h]
    \centering
    \includegraphics{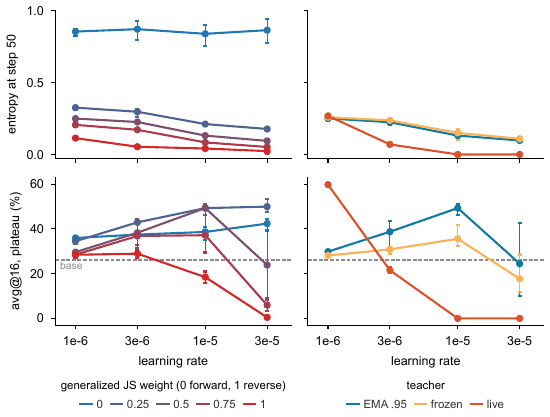}
    \caption{\textbf{The more a configuration deflates entropy, the smaller the learning rate it tolerates.}
    Entropy at step $50$ (top) and accuracy over the last five evaluations (bottom) against learning rate, varying the divergence weight (left) and the teacher (right), on GSM8K.
    Markers are means over three seeds and bars their range.
    JS stands for Jensen--Shannon and EMA for exponential moving average.}
    \label{fig:self-dist-lr}
\end{figure}

The learning rate sweep of Figure~\ref{fig:self-dist-lr} shows that the more a configuration deflates entropy, the smaller the learning rate it tolerates.
Reverse KL already loses to the base model at $10^{-5}$, whereas the arms closer to forward KL keep improving up to $3 \cdot 10^{-5}$, and a slowly moving or frozen teacher tolerates rates at which a live one has already collapsed.

\begin{figure}[h]
    \centering
    \includegraphics{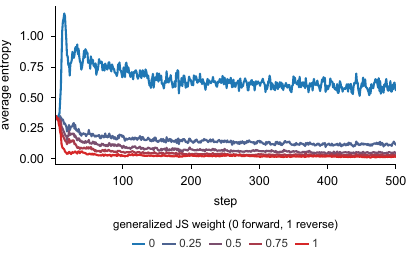}
    \caption{\textbf{Entropy settles within $500$ steps, and $\lambda=0$ is the only weight that ends up inflating it.}
    Entropy of the student over $500$ GSM8K steps, one seed per generalized Jensen--Shannon (JS) divergence weight.
    Forward KL converges at a high entropy level, every other $\lambda$ deflates below the initial entropy, and all configurations have flattened by the end of training, in the ordering the shorter runs already show.
    The gap between $\lambda=0$ and $\lambda=0.25$ is the largest and is consistent with a sharp phase transition in that interval, but the five weights are too coarsely spaced to confirm it.}
    \label{fig:self-dist-gsm-alpha-long}
\end{figure}

We finally use the same testbed to train ten times longer, and ask whether the sharp dependence of entropy on $\lambda$ that the toy model predicts at convergence (Figure~\ref{fig:toy_summary}d) appears once the runs are long enough.
Figure~\ref{fig:self-dist-gsm-alpha-long} shows that entropy does settle within $500$ steps.
The trajectories separate the two behaviors of Section~\ref{sec:entropy} cleanly: the forward KL arm has entropy much larger than its initial value, while every arm with $\lambda > 0$ settles below it.
The ordering in $\lambda$ holds at every point of the trajectory, and it leaves the gap between $\lambda = 0$ and $\lambda = 0.25$ as by far the largest of the sweep, as a phase transition inside that interval would predict.
Five weights spaced $0.25$ apart are too coarse to confirm it.
However, note that each run is evaluated on on-policy data, that is run-specific data, whereas the data was exactly the same in the toy setup.

\clearpage
\subsection{Hyperparameters of the self-distillation experiments}
\label{app:self-dist-hyperparameters}

\begin{table}[t]
    \centering
    \caption{\textbf{Default hyperparameters of the self-distillation experiments}, on the Chemistry dataset (Figure~\ref{fig:self-dist}) and on GSM8K (Figures~\ref{fig:self-dist-gsm} to \ref{fig:self-dist-gsm-alpha-long}).
    A value spanning both columns is shared by the two setups.
    Defaults follow the released implementation of \citet{hubotter2026rlsd}, and each ablation moves a single field away from them.
    The learning rate and its warmup are the exception: \citet{hubotter2026rlsd} report $10^{-6}$ without warmup for the setup we replicate and $10^{-5}$ with warmup for their other ones, and we use the latter, which reproduces the accuracy they report.}
    \label{tab:self-dist-hyperparameters}
    \vspace{0.3cm}
    \begin{tabular}{@{}lll@{}}
        \toprule
        & Chemistry & GSM8K \\
        \midrule
        \multicolumn{3}{@{}l}{\textit{Setup}} \\
        \midrule
        Trainer & \multicolumn{2}{l}{self-distillation trainer of TRL 1.9.2 \citep{vonwerra2020trl}} \\
        Reference implementation & \multicolumn{2}{l}{\citet{hubotter2026rlsd}} \\
        Sharding, generation engine & \multicolumn{2}{l}{DeepSpeed ZeRO-3, vLLM \citep{kwon2023vllm}} \\
        Dataset & SciKnowEval chemistry & GSM8K \\
        Student model & Olmo-3-7B-Instruct & Qwen3-0.6B \\
        Training prompts & $1890$ & $4000$ \\
        Thinking mode & \multicolumn{2}{l}{disabled} \\
        \midrule
        \multicolumn{3}{@{}l}{\textit{Rollouts}} \\
        \midrule
        Questions per optimizer step & \multicolumn{2}{l}{$32$} \\
        Rollouts per question & \multicolumn{2}{l}{$8$} \\
        Sampling temperature, top-$p$ & \multicolumn{2}{l}{$1.0$, $1.0$} \\
        Maximum prompt length & $2048$ & $512$ \\
        Maximum completion length & $2048$ & $640$ \\
        \midrule
        \multicolumn{3}{@{}l}{\textit{Objective}} \\
        \midrule
        Divergence & \multicolumn{2}{l}{generalized Jensen--Shannon, $\lambda = 0.5$} \\
        Token restriction & \multicolumn{2}{l}{top-$100$ with tail bucket} \\
        Importance ratio clip & \multicolumn{2}{l}{$2.0$, per token} \\
        Teacher & \multicolumn{2}{l}{moving average of the student, decay $0.95$} \\
        Privileged information & \multicolumn{2}{l}{successful rollout from the same group} \\
        Reference KL, reward scaling & \multicolumn{2}{l}{none} \\
        \midrule
        \multicolumn{3}{@{}l}{\textit{Optimization}} \\
        \midrule
        Optimizer & \multicolumn{2}{l}{AdamW, weight decay $0.01$, clipping $1.0$} \\
        Learning rate & $10^{-5}$ & $10^{-5}$ \\
        Schedule & \multicolumn{2}{l}{constant after $10$ warmup steps} \\
        Optimizer steps & $150$ & $50$ ($500$ in Figure~\ref{fig:self-dist-gsm-alpha-long}) \\
        Seeds & $1$ & $3$ ($1$ for the $500$-step runs) \\
        Precision & \multicolumn{2}{l}{bfloat16} \\
        Hardware & 8 H100 & 1 H100 \\
        \midrule
        \multicolumn{3}{@{}l}{\textit{Evaluation}} \\
        \midrule
        Evaluation sampling & \multicolumn{2}{l}{temperature $0.6$, top-$p$ $0.95$} \\
        Evaluation questions & $210$ & $200$ held out \\
        Evaluated every & $50$ steps & $5$ steps \\
        \bottomrule
    \end{tabular}
\end{table}

\end{document}